\documentclass[]{assets/template}
\usepackage{hyperref}
\usepackage{url}
\usepackage{amsthm}
\usepackage{fontawesome5}
\usepackage[smartEllipses]{markdown}
\usepackage[most]{tcolorbox}
\usepackage{wrapfig}
\usepackage{multirow}
\usepackage{makecell}
\usepackage{graphicx}
\usepackage{threeparttable}
\usepackage[ruled,vlined]{algorithm2e}
\usepackage{colortbl}
\usepackage{color}
\usepackage{multirow}
\usepackage{tabularx}
\usepackage{float}
\usepackage{graphicx}
\usepackage{booktabs}
\usepackage{arydshln}
\usepackage{enumitem}
\usepackage{wrapfig}
\usepackage{caption}
\usepackage{graphicx}
\usepackage{makecell}
\usepackage{float} 
\usepackage{mathtools}
\usepackage{bbding}
\usepackage{makecell}
\usepackage{tabularx}
\usepackage{amssymb,mathrsfs,amsmath}
\usepackage{pifont}
\usepackage{xcolor}
\usepackage{bbm}
\usepackage{datetime}
\usepackage{bxcoloremoji}
\usepackage{listings}
\usepackage{subcaption}
\usepackage{cleveref}
\usepackage{calc}
\usepackage{bxcoloremoji}         
\usetikzlibrary{calc}
\usepackage{hyperref}
\addtocontents{toc}{\protect\setcounter{tocdepth}{0}}

\newdateformat{usvardate}{\monthname[\THEMONTH] \space \THEDAY, \space \THEYEAR}

\newcommand{\fucla}{\scalebox{0.67}{$\bigcirc$}}
\newcommand{\fms}{\square} 
 
\newcommand{\feth}{\triangle}

\newcommand{\ourb}{HAGeo-409}
\newcommand{\ourm}{HAGeo}

\newlength{\BWidth}    
\newlength{\BRadius}   
\newlength{\LRadius}   
\newlength{\LSep}      
\newlength{\LShift}    
\newlength{\BLine}     
\newlength{\BannerSep} 
\newcommand{\BFont}{\small}
\newcommand{\LFont}{\small}
\newcommand{\BannerThemeAnswer}{
  \def\BFill{gray!5!white}
  \def\BDraw{gray!60!black}
  \def\LFill{gray!10}
  \def\LDraw{gray!50!black}
}
\definecolor{metagreen}{HTML}{2E8B57} 
\definecolor{msblue}{RGB}{0,102,204}
\definecolor{heuristic1}{RGB}{205,205,205}
\definecolor{heuristic2}{RGB}{255,155,76}
\definecolor{heuristic3}{RGB}{60,135,72}
\definecolor{heuristic4}{RGB}{3,255,254}
\definecolor{heuristic5}{RGB}{158,75,249}
\definecolor{heuristic6}{RGB}{255,77,249}

\title{Gold-Medal-Level Olympiad Geometry Solving with Efficient Heuristic Auxiliary Constructions}

\author[~\feth]{Boyan Duan}
\author[~\fucla\;\dagger]{Xiao Liang}
\author[~\fms]{Shuai Lu}
\author[~\fms]{Yaoxiang Wang}
\author[~\fms]{Yelong Shen}
\author[~\fucla]{Kai-Wei Chang}
\author[~\fucla]{\\Ying Nian Wu}
\author[~\fms]{Mao Yang}
\author[~\fms]{Weizhu Chen}
\author[~\fms\;\dagger]{Yeyun Gong}

\affiliation[\feth]{ETH Zurich}
\affiliation[\fms]{Microsoft}
\affiliation[\fucla]{University of California, Los Angeles}
\contribution[]{Work done during Boyan, Xiao, and Yaoxiang's intership at Microsoft}
\contribution[\dagger]{Corresponding authors.}

\begin{document}
\abstract{
Automated theorem proving in Euclidean geometry, particularly for International Mathematical Olympiad (IMO) level problems, remains a major challenge and an important research focus in Artificial Intelligence.
In this paper, we present a highly efficient method for geometry theorem proving that runs entirely on CPUs without relying on neural network–based inference.
Our initial study shows that a simple random strategy for adding auxiliary points can achieve ``silver-medal'' level human performance on IMO. 
Building on this, we propose \textbf{\ourm}, a \underline{H}euristic-based method for adding \underline{A}uxiliary constructions in \underline{Geo}metric deduction that solves \textbf{28 of 30} problems on the IMO-30 benchmark, achieving ``gold-medal'' level performance and surpassing AlphaGeometry, a competitive neural network–based approach, by a notable margin.
To evaluate our method and existing approaches more comprehensively, we further construct \textbf{\ourb}, a benchmark consisting of 409 geometry problems with human-assessed difficulty levels. 
Compared with the widely used IMO-30, our benchmark poses greater challenges and provides a more precise evaluation, setting a higher bar for geometry theorem proving. \\
\\
\coloremojicode{1F4C5}~ \textbf{Date}: \usvardate\today \\
\faGithub~ \textbf{Code}: \href{https://github.com/boduan1/HAGeo}{https://github.com/boduan1/HAGeo} \\
\coloremojicode{1F917}~ \textbf{Benchmark}: \href{https://huggingface.co/datasets/HAGeo-IMO/HAGeo-409}{https://huggingface.co/datasets/HAGeo-409} \\
\coloremojicode{2709}~ \textbf{Email}: Xiao Liang (\href{vitoliang0601@gmail.com}{vitoliang0601@gmail.com}); Yeyun Gong (\href{yegong@microsoft.com}{yegong@microsoft.com})
\vspace{-10pt}
}
\maketitle
\section{Introduction}
\label{sec:introduction}
Automatic mathematical theorem proving is one of the fundamental goals of Artificial Intelligence (AI). 
As a branch of mathematics studied for over two millennia, Euclidean geometry serves as one of the four primary problem categories in the International Mathematical Olympiad (\href{https://www.imo.org/}{IMO}), the world’s leading high school mathematics competition.
Given its enduring importance, automated theorem proving in Euclidean geometry has remained a focus of research and has been extensively studied for decades~\citep{gelernter1959realization,gelernter1960empirical,reiter1972use}.

As a recent advance, AlphaGeometry \citep{trinh2024solving} demonstrates remarkable progress by utilizing a deduction database and an algebraic reasoning (DDAR) engine.
It employs a neural–symbolic system to exhaustively deduce new statements using a symbolic engine and a neural language model trained on hundreds of millions of synthetic data to add auxiliary points for Euclidean geometry problem solving. 
It achieves a ``silver-medal'' performance on the IMO-30 benchmark, a collection of 30 competition-level geometry problems from the International Mathematical Olympiad between 2000 and 2022, solving 24 of the 30 problems \footnote{\noindent We report that one of the 25 announced proofs by AlphaGeometry on IMO-30 is wrong in Appendix~\ref{sec:AlphaGeometry-false-positive}.}. 
More recently, by incorporating multiple Large Language Models (LLMs) instead of a small neural network and introducing a shared knowledge mechanism,  AlphaGeometry2~\citep{chervonyi2025gold} successfully solves all problems in IMO-30.
Meanwhile, TongGeometry~\citep{zhang2024proposing} achieves the same performance by employing neural models with guided tree search; however, its technical details remain unreported.

Despite their effectiveness, a common limitation of these methods is the reliance on neural models, particularly LLMs, which requires additional GPU resources for inference.
In our initial study, we find that even a simple random strategy for adding auxiliary points using only CPUs allows the system to reach AlphaGeometry's performance on IMO-30, solving 25 of 30 problems.
These surprising results raise a promising research question: \textit{Could the system achieve ``gold-medal'' performance by simply employing an efficient strategy for adding auxiliary points without relying on neural inference?}

To answer this, we propose \textbf{\ourm}, a heuristic method that relies solely on adding auxiliary constructions in deduction without using a neural network, outperforming AlphaGeometry and achieving ``gold-medal'' performance on the IMO-30 benchmark.
Specifically, \ourm~adds heuristic auxiliary points with favorable geometric properties, such as the intersections of circles and lines, which can be discovered through numerical calculations. 
Given that the deduction engine already assumes all numerical information is available, our method introduces no additional assumptions about numerical values.
Moreover, we develop optimized deduction rules to lower time complexity without sacrificing deductive capability and optimize the implementation, yielding an approximately $20\times$ speedup over AlphaGeometry's DDAR engine. These improvements enable faster and more scalable inference.

For evaluation, while the IMO-30 benchmark is widely used, we find that it overlooks problem difficulty and most of its problems are relatively easy, as judged by professionals.
Besides, it consists of only 30 problems, which could lead to high variance in evaluation.
Therefore, this benchmark does not reliably reflect a method’s true capability.
To address the limitations of IMO-30 and more comprehensively evaluate geometry problem solving, we construct \textbf{\ourb}, a benchmark consisting of 409 Olympiad-level geometry problems with human-assessed difficulty labels and generally more challenging than IMO-30.
The problems have been systematically converted into geometry-specific language assisted by LLMs, numerically verified, and corrected through manual inspection, making them more rigorous for evaluation.

\noindent \textit{In summary, our main \textbf{contributions} are as follows:}
\begin{itemize}[leftmargin=0.95em, topsep=0.5pt, partopsep=0pt, itemsep=0.5pt, parsep=0pt]

\item We propose \ourm, a heuristic method for adding auxiliary points in automated theorem proving for Euclidean geometry, achieving ``gold-medal'' performance on CPUs and surpassing AlphaGeometry's neural network–based approach.

\item We implemented an improved DDAR engine that achieves roughly a 20× inference speedup compared to AlphaGeometry's DDAR engine. 

\item We construct \ourb, a comprehensive benchmark comprising 409 IMO-level geometry problems with human-assessed difficulty scores that are generally more challenging than those in IMO-30. Experiments on both benchmarks demonstrate the effectiveness of our method.
\end{itemize}

\section{Related Work}
\subsection{Automated Theorem Proving for Euclidean Geometry}

Existing methods for automated theorem proving in Euclidean geometry can be broadly divided into algebraic and synthetic approaches.
Algebraic methods typically transform a geometry problem into a system of polynomial equations, which can then be solved using Wu's method \citep{wu1978decision,chou1988introduction} or the Gröbner basis method \citep{lazard1983grobner}.
For synthetic methods, \citet{chou1993automated} applied the area method and generated human-readable proofs for more than 400 geometry problems, while \citet{chou2000deductive} introduced a deductive database (DD) based on a set of geometric rules. 

Building on synthetic methods, remarkable progress has recently been made in the field. 
AlphaGeometry \citep{trinh2024solving} improves the \textit{DD method with an additional algebraic engine} (DDAR) and employs a neural network to add extra auxiliary points for geometry problems. It achieves a ``silver-medal'' performance on the IMO-30 benchmark.
When combined with Wu's method \citep{sinha2024wu}, which independently solves 15 of the 30 problems on IMO-30, AlphaGeometry can solve 27/30 problems.
More recently, TongGeometry \citep{zhang2024proposing} improved AlphaGeometry's DD engine and proposed a tree-search-based method that solves 30/30 problems on IMO-30, but the technical details remain unavailable.
AlphaGeometry2 \citep{chervonyi2025gold} introduces several improvements over AlphaGeometry, including enhancements to its geometric language, increased efficiency of the DDAR engine, and support for the ``double point'' feature in DDAR.
It further extends the neural network with an ensemble of multiple language models and introduces shareable knowledge across different searches, while also refining the synthetic dataset with more complex problems and a better-balanced distribution.

\subsection{Geometry Problem Dataset}
Prior work has collected multiple Olympiad-level mathematics datasets for theorem proving~\citep{ying2024lean,xin2024deepseek,wei2024proving,de2015lean,zheng2021minif2f}.
However, these datasets primarily focus on algebra and number theory in Olympiad competitions, with only a few of them dedicated to Olympiad-level geometry theorem proving. 
Furthermore, most of them are based on \textit{Lean}~\citep{moura2021lean}, which is less effective for geometry problems.
Recent work~\citep{liang2025sws,li2025system} shows even frontier LLMs exhibit weaker performance on geometry than on algebra, and data augmentation targeted at this challenging subdomain has shown promise in mitigating this issue~\citep{liang2025beyond}.

Recently, AlphaGeometry generated a large-scale synthetic dataset with 100 million problems to train a neural network for adding auxiliary points. 
It randomly generates geometric configurations and uses the DDAR engine to deduce geometric facts, which are then considered as geometry problems.
GeoGen~\citep{bak2020automated} proposes adding a few objects to triangle and quadrilateral configurations to extract geometry theorems, generating more than 100 turnaround problems.
Using a method similar to GeoGen, TongGeometry generates over 100 million geometric configurations and extracts 6 billion geometry problems.
However, their further technical details are not provided. 

\section{\ourb~Benchmark Construction}
Previous studies, such as AlphaGeometry, primarily employed two benchmarks to evaluate their Euclidean geometry theorem-proving systems: JGEX-231~\citep{JGEX231} and IMO-30. 
The JGEX-231 benchmark consists of 231 relatively simple geometry problems, while IMO-30 includes only 30 problems drawn from International Mathematical Olympiads spanning 2000 to 2022.
Furthermore, as in the IMO competition, problems 1 and 4, 2 and 5, and 3 and 6 typically differ substantially in difficulty; however, we observe that the IMO-30 benchmark predominantly contains problems that are not especially challenging.

Since it is important to know the upper limit of problem difficulty that a method could solve, in this work, we constructed a more comprehensive benchmark from Mathematical Olympiad competitions, with each problem annotated with human-evaluated difficulty levels.
We first collected more than 2000 
geometry problems from the contest data collection of Art of Problem Solving website~\citep{aops}. 
Then, we converted these problems into our geometry-specific language, with an illustration provided in Appendix~\ref{sec:example}. 
The successfully converted raw dataset contained more than 1,000 geometry problems, all of which were verified numerically, ensuring that each problem’s conclusion is numerically correct.

\begin{figure*}[t]
    \centering
    \includegraphics[width=\linewidth]
    {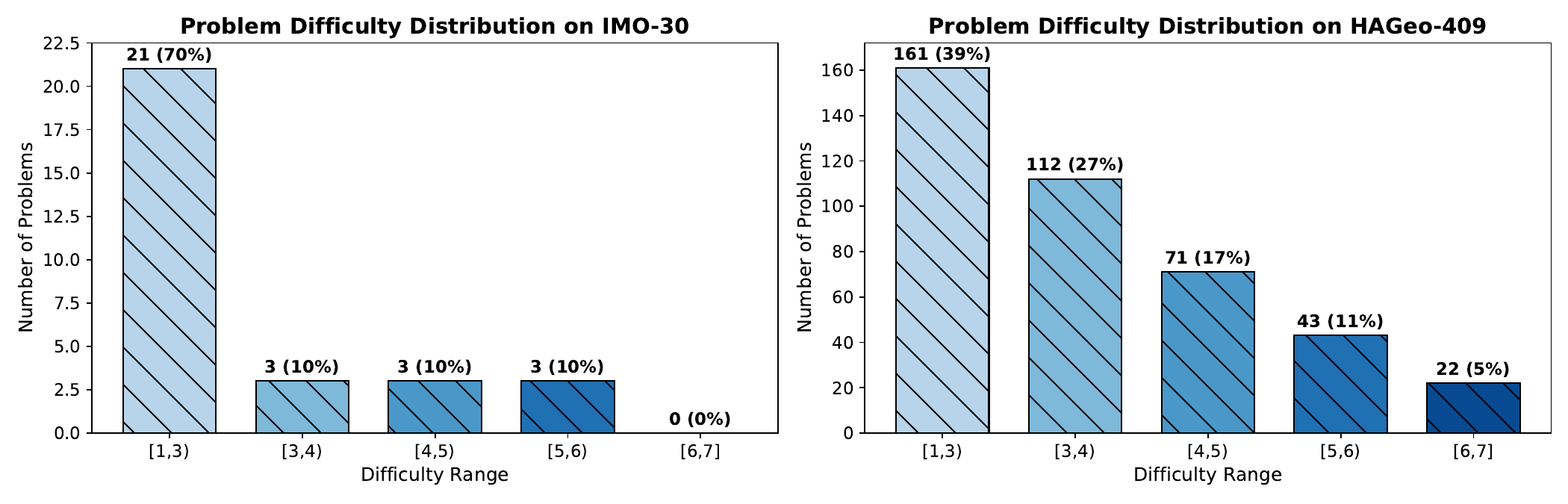}
    \vspace{-1.5em}
    \captionsetup{justification=centering}
    \caption{Problem difficulity distribution of the IMO-30 benchmark and our new \ourb~benchmark.}
    \vspace{-15pt}
    \label{fig:benchmark}
\end{figure*}

However, not all problems in the raw dataset are paired with a human-annotated difficulty score. 
We retain a subset of problems, along with around 50 additional problems from the mathematical platform ShuZhiMi\footnote{A \href{https://www.wechat.com/}{Wechat} mini program dedicated to Olympiad-level mathematical problems.}. 
The difficulty level ranges from 1 to 7 (corresponding to easy to hard) and is defined as the average rating provided by ShuZhiMi users. 
The final benchmark contains over 400 problems with difficulty level annotations. 
We categorize the benchmark into difficulty ranges $[1,3), [3,4), [4,5), [5,6), [6,7]$ and present the statistics in Figure~\ref{fig:benchmark}.
Notably, compared with \ourb, whose average difficulty is 3.47, the IMO-30 benchmark is relatively easy, with an average difficulty of only 2.85.

We also report that the conversion process is non-trivial. 
We employ GPT-4o~\citep{hurst2024gpt} with a few-shot prompt, as illustrated in Prompt~\ref{prompt:sigmageometry},
to convert geometry problems from natural language into our geometry-specific language. 
However, only around $50\%$ of the problems could be converted into a construction-based definition, and less than $20\%$ could be automatically converted and numerically verified. 
We then manually revised the remaining problems that could be expressed in our geometric language and obtained the raw dataset. 
To evaluate AlphaGeometry on our \ourb, we used GPT-4o with Prompt~\ref{prompt:alphageometry} to convert the problems into the AlphaGeometry format, and manually corrected errors in the converted problems.
To ensure a fair comparison with AlphaGeometry, we further refined \ourb~to align with the its format as closely as possible.
For example, if AlphaGeometry defines the circumcircle as \textit{ o = circumcenter a b c; d = oncircle o a; } (which introduces the circumcenter), we also add this point in our setting.

\begin{figure*}
  \centering
    \includegraphics[width=1.\linewidth]{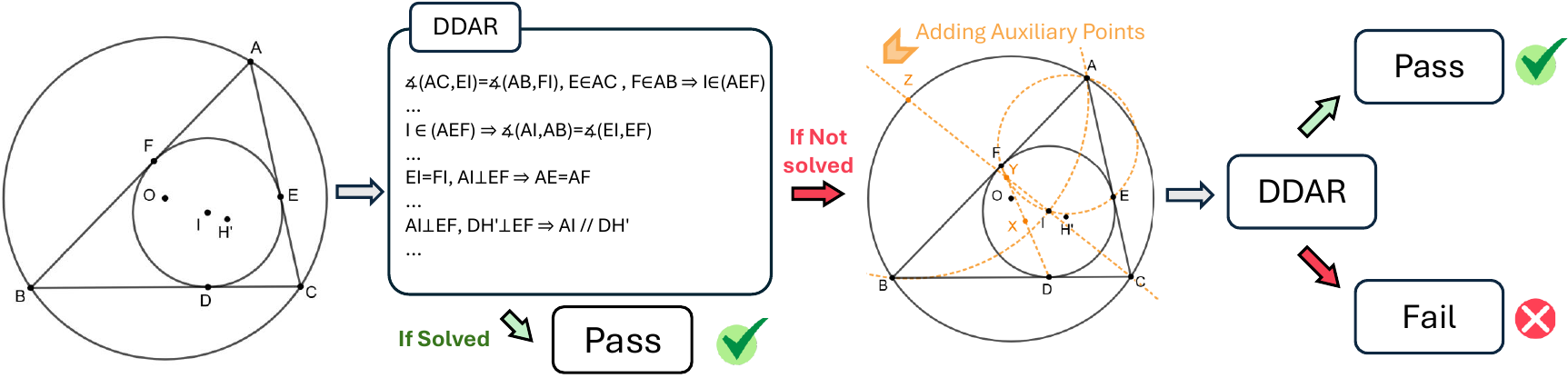}
    \vspace{-10pt}
    \caption{Overview of the \ourm~method. First, the DDAR engine deduces new statements in the problem. If the DDAR does not solve the problem, our heuristic-based strategy gives additional attempts for adds auxiliary constructions to help solve the problem and re-runs the DDAR.}
    \vspace{-15pt}
    \label{fig:overview}
\end{figure*}

\section{Method}
\vspace{-5pt}
\subsection{General Pipeline}
\vspace{-3pt}
In \ourm, a geometry problem is first converted into our geometry-specific language and then processed to encode all its properties into a deduction graph.
The Deduction Database (DD) (Section~\ref{sec:deduction}) and Algebraic Reasoning (AR) (Section~\ref{sec:algebraic-reasoning}) engines expand the deduction graph through a brute-force search over all deduction rules and by performing algebraic deductions involving equations of length, ratio, or angle.
If the DDAR fails to solve the problem, we further attempt $K$ different auxiliary constructions based on our heuristic strategies, and the DDAR engine is rerun for each attempt, as detailed in Section~\ref{sec:heuristic}.
Additional details on DDAR are provided in Appendix~\ref{sec:ddar-preliminary}.
The overview of \ourm~is shown in Figure~\ref{fig:overview}.

\begin{figure*}[t]
  \centering
    \includegraphics[width=0.9\linewidth]{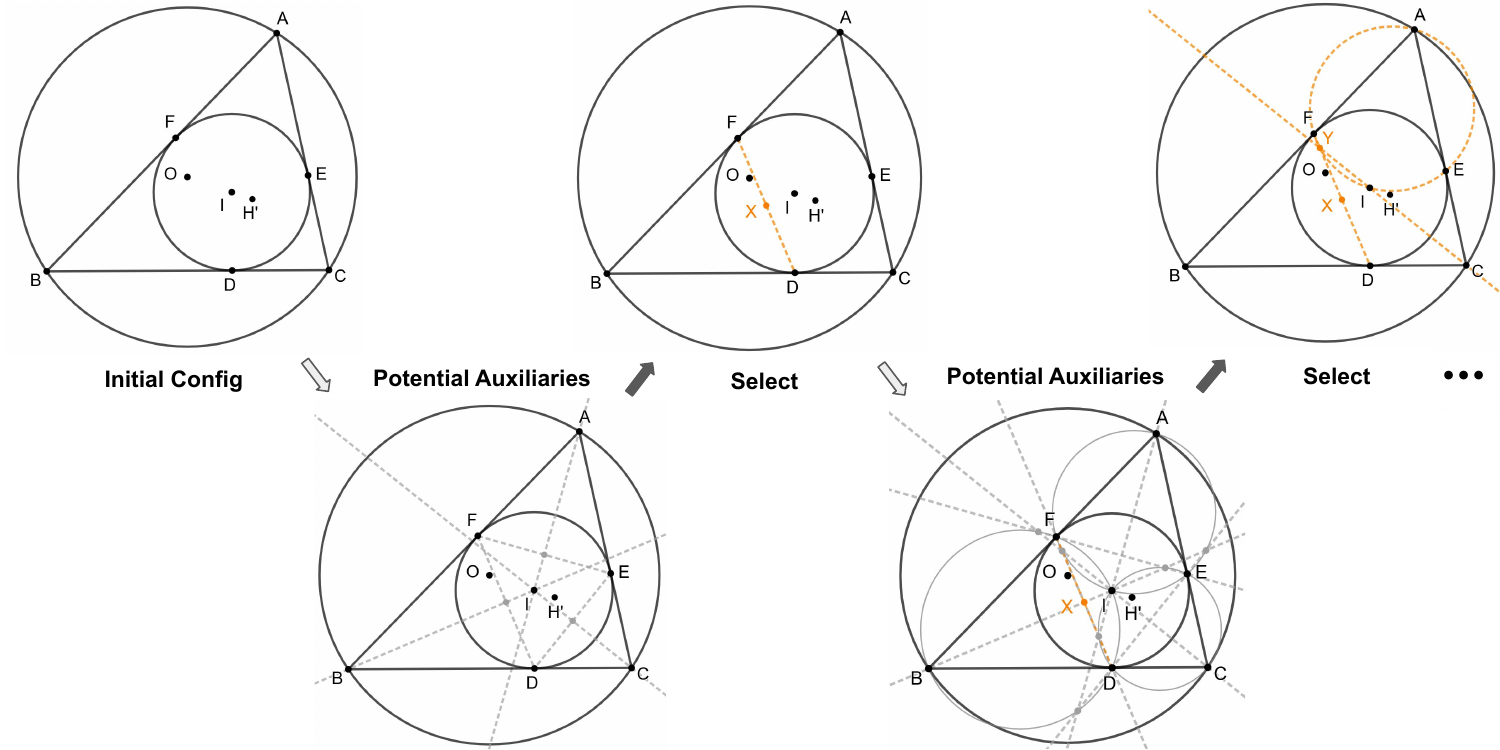}
    \vspace{-5pt}
    \caption{Pipeline for adding heuristic auxiliary points. Each complete auxiliary construction consists of up to $N$ rounds of auxiliary point generation. In each round, the system selects one valid auxiliary point from all possible candidates determined through algebraical computation.}
    \vspace{-5pt}
    \label{fig:aux}
\end{figure*}

\subsection{Geometry-Specific Languages}
\vspace{-3pt}

Unlike theorem proving in other mathematical domains, such as algebra, which can be formally expressed in \textit{Lean}, geometry lacks a standardized formal language for theorem proving.
In \ourm, we adopt the geometry-specific language in GeoGebra~\citep{geogebra}, which includes definitions of points, lines, and circles, along with a fixed set of construction actions.

For example, consider a problem that defines new points $X,Y$ as the intersection of line $AB$ and a circle that is centered at $O$ and passes through point $P$. 
Our geometric language first defines the line $AB$ : 
\texttt{l = line A B}, then circle $(O,OP)$: \texttt{$\omega$  = circle\_center\_point O P}, and then define the points $X,Y$: \texttt{X, Y = intersection l $\omega$}.
When the problem involves more advanced constraints, we address them through curve intersections. For example, if the problem defines a point $X$ that satisfies $\angle AXB = \angle CDE$ and $X$ lies on line $PQ$, we first define a curve $\omega$ with the first condition, then set $X$ as the intersection of $\omega$ and line $PQ$.

In contrast to AlphaGeometry, which employs a point-based geometric language that considers only points, 
our definition better reflects the natural language of problem statements, encompassing diverse geometric objects.
This design is also more consistent with the deduction engine that includes rules with multiple objects, such as lines and circles, rather than only points as in AlphaGeometry.

\subsection{Deduction}
\label{sec:deduction}
The symbolic deduction engine deduces new statements by brute-force searching over a fixed set of deduction rules. 
Each deduction rule has the form $P_1(x_1),\dots,P_k(x_k) \Rightarrow Q(x)$, where $P_1,...P_k$ and $Q$ are predicates, and $x_1,\cdots,x_k$ and each $x$ represent a set of geometric objects. 
Since defining the same point in different ways is an important technique, we extend the DD engine to support deductions involving identical points, whereas AlphaGeometry’s engine restricts all points to be distinct.

The deduction is represented as a graph, where geometric objects and relations are represented as vertices and edges.
Whenever a new property is deduced, the deduction graph is updated accordingly—by adding vertices or edges, or by merging existing ones.
To enhance scalability and inference speed, we develop a deduction engine that operates with significantly higher efficiency, \textit{running approximately 20 times faster than AlphaGeometry.}
Specifically, we evaluate both methods on the same machine. The average solving time of AlphaGeometry’s DDAR on IMO-30 is $42.77$ seconds, whereas ours achieves an average of $1.75$ seconds, representing a 24× speed improvement.

The improved inference efficiency stems from the modifications to the deduction rules and the optimizations in their implementation.
We provide two examples in the following grey box.
These updated rules reduce time complexity while preserving the same deductive capacity, thereby improving the efficiency of most time-consuming rules and significantly accelerating the speed of the deduction engine.
\begin{figure}[ht]\footnotesize
\vspace{-5pt}
\begin{tcolorbox}[colback=gray!5!white, colframe=gray!60!black, boxsep=2pt, left=2pt, right=2pt, top=1.5pt, bottom=1.5pt]
1. We replace (angle-chasing rule on lines $l_i,m_i$):
\begin{center}
\textcolor{blue}{
$\measuredangle (l_1,l_2) = \measuredangle (m_1,m_2),\;
\measuredangle(l_1,l_3) = \measuredangle(m_1,m_3)$\\
$\Rightarrow \measuredangle (l_2,l_3) = \measuredangle (m_2,m_3)$
}\\
$\Downarrow$ \\
\textcolor{green}{$\measuredangle (l_1,l_2) = \measuredangle (m_1,m_2) \Rightarrow \measuredangle (l_1,m_1) = \measuredangle (l_2,m_2) $}
\end{center}
\vspace{\baselineskip}
(2) We replace (positive similar triangle rule on $\Delta ABC, \Delta DEF$):
\begin{center}
\textcolor{blue}{
$\measuredangle (AB,BC) = \measuredangle (DE,EF), \measuredangle (AC,BC) = \measuredangle (DF,EF) \Rightarrow \Delta ABC \stackrel{+}{\sim} \Delta DEF$}\\
$\Downarrow$ \\
\textcolor{green}{$\measuredangle (AB,DE) = \measuredangle (AC,DF) = \measuredangle (BC,EF)$} \\ \textcolor{green}{$\Rightarrow \Delta ABC \stackrel{+}{\sim} \Delta DEF$}
\end{center}
\end{tcolorbox}
\vspace{-5pt}
\end{figure}

\subsection{Algebraic Reasoning}
\label{sec:algebraic-reasoning}

Building on the geometric properties deduced by the DD engine, we further adapt an AR engine to infer additional statements algebraically.
The AR engine first converts all length, ratio, and angle relations into linear equations. For example, it converts $\measuredangle (l_1,l_2) = \measuredangle(l_3,l_4)$ into $\texttt{dir}(l_1)+\texttt{dir}(l_4)-\texttt{dir}(l_2)-\texttt{dir}(l_3) = 0$, where $\texttt{dir}(l)$ represents the direction of the line as the directed angle of $l$ with the $x$ axis $mod\ \pi$.
It then aggregates all linear equations into a coefficient matrix and derives new relations accordingly.
Specifically, it uses Gaussian Elimination~\footnote{\href{https://en.wikipedia.org/wiki/Gaussian_elimination}{en.wikipedia.org/wiki/Gaussian\_elimination}} to identify the independent set of variables and express all variables as linear combinations of this set.
It then expresses all $x_i - x_j$ as linear combinations of the independent variables and finds all equivalent $x_{i1} - x_{j1} = x_{i2} - x_{j2}$.

It is worth noting that our implementation is fully based on matrix manipulation.
To reduce matrix size and improve computational efficiency, we merge all equivalent variables before constructing the matrix.
For the final step, we use an equivalent alternative implementation that takes all $x_i + x_j$ instead of $x_i - x_j$ and finds all $x_{i1} + x_{j1} = x_{i2} + x_{j2}$, which halves the computation due to the symmetry.

\subsection{Heuristic for Auxiliary Points}
\label{sec:heuristic}

Adding auxiliary constructions is an important strategy for solving geometry problems and can be regarded as selecting from the set of all valid auxiliary objects.
However, while the number of valid constructions can be very large, only a small set of them is useful.
AlphaGeometry employs a neural network to construct auxiliary points while reporting that its implemented heuristic approach solves only 18 out of 30 problems on the IMO-30 benchmark. 
However, in their implementation, the rule-based heuristic only uses fixed rules to add auxiliary points. 
For example, \textit{"If $M$ is the midpoint of $AB$, and $N$ is the midpoint of $CD$, then add the midpoint $P$ of $AD$ as an auxiliary point"}.

\begin{figure}[t]
    \centering
    \begin{tikzpicture}
      \BannerThemeAnswer
      \coordinate (C) at (\dimexpr.5\linewidth\relax,0);
    
      \node[
        draw=\BDraw, fill=\BFill,
        rounded corners=\BRadius, line width=\BLine,
        minimum width=\BWidth,
        text width=\dimexpr\BWidth - \BLine - 1\BannerSep\relax,
        inner sep=\BannerSep,
        align=left,
        font=\BFont
      ] (main) at (C) {
        \\
        \textbf{1. The intersection of multiple lines.} \\
        \textit{If multiple lines $l_1, \dots, l_n$ intersect ($n \ge 3$), then take the intersection of two of them as an auxiliary point.}\\
        Eg.1: \textcolor{blue}{The lines $AD,BE,CF$ intersects, and we can take $U = AD \cap BE$ as an auxiliary point.}\\
        \vspace{\baselineskip}
        
        \textbf{2. The intersection of multiple lines and circles.} \\
        \textit{If multiple lines $l_1, \dots, l_n$ and circles $\omega_1, \dots, \omega_m$ intersect ($m+n \ge 3, n \ge 1$), then take the intersection of a line and a circle ($m \ge 1$) or two circles ($n \ge 3$) as an auxiliary point.}\\
        Eg.2: \textcolor{heuristic2}{The lines $AI,OL$ and circle $(ABC)$ intersects, and we can take $V$, the intersection of $AI$ and $(ABC)$ as an auxiliary point.}\\
        \vspace{\baselineskip}
        
        \textbf{3. A midpoint non-trivially lies on lines or circles.} \\
        \textit{If the midpoint of $AB$ lies on a line $l \neq AB$ or a circle $\omega$, then take it as an auxiliary point.}\\
        Eg.3: \textcolor{heuristic3}{The midpoint of $AH$, defined as $W$, lies on circle $(LMN)$, and we take $W$ as an auxiliary point.}\\
        
        \vspace{\baselineskip}
        \textbf{4. Reflection of a point with respect to another point non-trivially lies on lines or circles.} \\
        \textit{If the reflection of a point $A$ wrt another point $B$ lines on line $l \neq AB$ or circle $\omega$, then take it as an auxiliary point.} \\
        Eg.4: \textcolor{heuristic4}{The reflection of $H$ wrt $L$, defined as $X$, lies on circle $(ABC)$, and we take $X$ as an auxiliary point.}\\
        \vspace{\baselineskip}
        
        \textbf{5. Foot of a point lies on another line.} \\
        \textit{If the foot of a point $A$ on line $l$ ($A \notin l$) lies on another line $m$ ($m \neq l, A \notin m$), then take it as an auxiliary point.}\\
        Eg.5: \textcolor{heuristic5}{The foot of point $B$ on line $AI$ lies on line $LN$, defined as $Y$, and we take $Y$ as an auxiliary point.}\\
        \vspace{\baselineskip}
        
        \textbf{6. Random constructions.} \\
        \textit{Randomly perform an action over a random set of geometry objects corresponding to the action.}\\
        Eg.6: \textcolor{heuristic6}{Construct a point $Z$ such that $\Delta ZHD$ is a clockwise equilateral triangle.}
      };
    
      \node[
        draw=\LDraw, fill=\LFill,
        rounded corners=\LRadius, inner sep=\LSep,
         anchor=west,
        line width=\BLine, font=\LFont
      ] at ($(main.north west)+(\LShift+1,0)$) {\textbf{Heuristics for Auxiliary Points}};
    
    \end{tikzpicture}
    \vspace{-15pt}
    \caption{The heuristics proposed in \ourm. The paired visualization is present in Figure~\ref{fig:aux_example}. Each heuristic is exemplified using a distinct color, which are aligned with the lines and points shown in Figure~\ref{fig:aux_example}.}
    \vspace{-10pt}
    \label{fig:heuristics}
\end{figure}

\begin{wrapfigure}{r}{0.4\textwidth}
    \centering
    \vspace{-1.05em}
    \includegraphics[width=\linewidth]{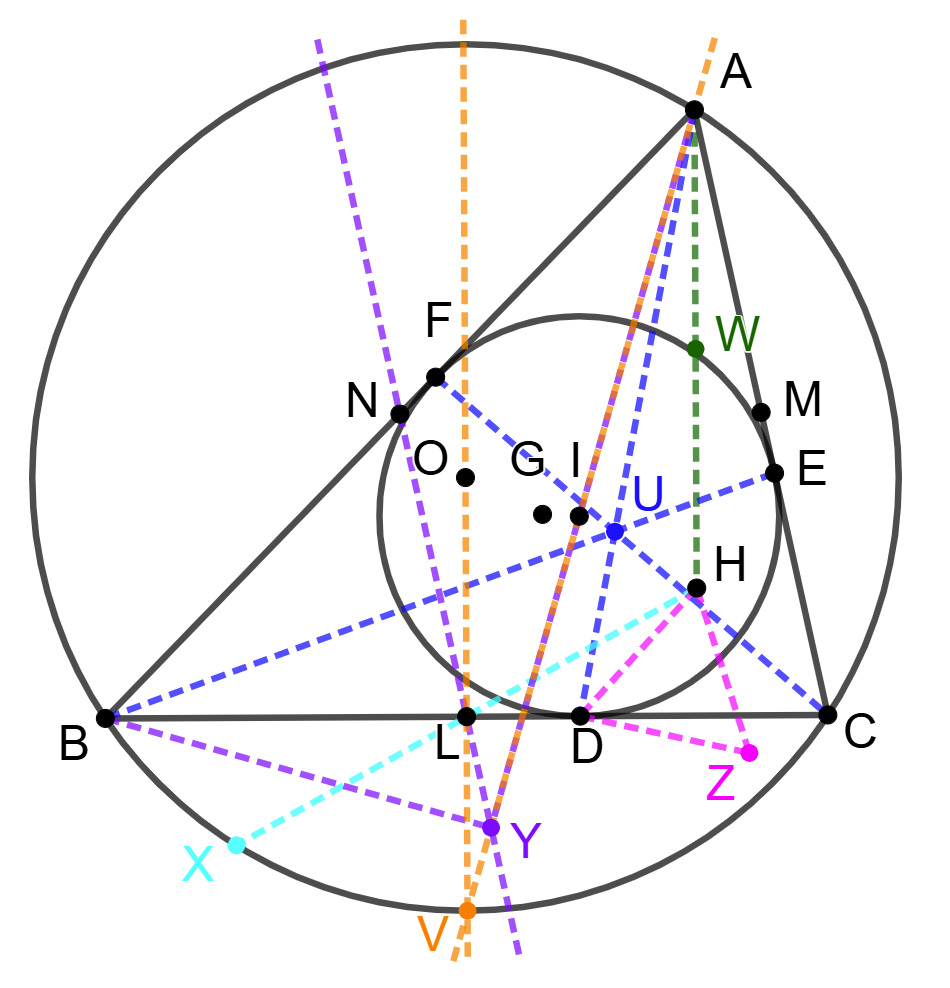}
    \vspace{-0.98em}
    \caption{\textbf{Illustration of heuristic auxiliary points.}
    Geometry configuration: $I,G,O,H$ are the incenter, centroid, circumcenter, and orthocenter of $\triangle ABC$; $L,M,N$ are the midpoints of $BC,CA,AB$; and $D,E,F$ are the tangent points of the incircle $(I)$ with sides $BC,CA,AB$.}
    \label{fig:aux_example}
    \vspace{-2.6em}
\end{wrapfigure}
In geometry, we usually add auxiliary constructions with good geometric properties; for example, the intersection of multiple curves may yield a promising auxiliary point. 
Guided by this principle, our heuristic designates a constructed point as a potential auxiliary point if it non-trivially lies on a line or circle.

Specifically, we employ a numerically driven auxiliary-point heuristic that only includes basic auxiliary constructions, such as the midpoint, reflection, foot of the perpendicular, and intersection, since it is easy to verify whether the resulting relation is trivial. 
Importantly, the numerical calculations introduce no extra assumptions for assigning numerical values to all geometric objects. 
This is because the deduction step (as detailed in Section~\ref{sec:deduction}) already requires numerical checks, such as verifying that points $A,B,C$ are not collinear, which inherently assumes that numerical values for all geometric objects are available.

We categorize auxiliary points into several categories, as detailed in Figure~\ref{fig:heuristics} and illustrated in the paired visualization in Figure~\ref{fig:aux_example}, including previously mentioned points with favorable geometric properties such as \textit{intersections}, \textit{midpoints}, and \textit{reflections}.
The random auxiliary points are also incorporated to enhance the generalization of the procedure.
Additionally, we also provide a special heuristic for ``identical points'': if an existing point lies on multiple circles and lines, this point of intersection (which is identical to the existing one) is also selected as an auxiliary point.

If the system fails to solve a geometric problem in the initial DDAR, we add auxiliary points based on our heuristics to assist the DDAR engine for up to $K$ additional attempts.
Specifically, during each auxiliary construction attempt, the system first calculates all potential auxiliary points from the aforementioned heuristic categories and then randomly selects one point(s) along with its corresponding construction.
It repeats this process for $N$ rounds and obtains the final set of auxiliary constructions, which are integrated into the new DDAR runs.
It is worth noting that both the DDAR and heuristic auxiliary construction procedures are fully executed on CPUs without neural inference.
The experiments in Section~\ref{sec:results} show that equipping DDAR with our proposed heuristic auxiliary adding approach yields high efficiency and effectiveness, surpassing both the random selection baseline and the neural-network-based strategy in AlphaGeometry.
\section{Experiments}

\begin{table*}[t]
  \centering
    \resizebox{1.0\textwidth}{!}{%
    \footnotesize
    \begin{tabular}{lcccccc}
        \toprule[1.0pt]
        \toprule[1.0pt]
        Level & Count & \makecell{AlphaGeometry\\16-64-8}  & \makecell{Random\\@2048} & \makecell{\ourm\\@2048} & \makecell{Random\\@8192} & \makecell{\ourm\\@8192} \\
        \midrule
        1--3   & 161 & 118 (73.3\%) & 127 (78.9\%) & 141 (87.6\%) & 128 (79.5\%) & 149 (92.5\%) \\
        3--4   & 112 & 44 (39.3\%)  & 62 (55.4\%)  & 87 (77.7\%)  & 69 (61.6\%) & 93 (83.0\%) \\
        4--5   & 71 & 13 (18.3\%)   & 13 (18.3\%)   & 29 (40.8\%)   & 18 (25.4\%)   & 36 (50.7\%)  \\
        5--6   & 43 & 2 (4.7\%)    & 2 (4.7\%)    & 5 (11.6\%)    & 3 (7.0\%)    & 7 (16.3\%)  \\
        6--7   & 22 & 0 (0.0\%)    & 0 (0.0\%)    & 1 (4.5\%)    & 0 (0.0\%)    & 2 (9.1\%)  \\
        \midrule
        Total  & 409 & 177 (43.3\%) & 204 (49.9\%) & 263 (64.3\%) & 218 (53.3\%) & 287 (70.2\%) \\
        \bottomrule[1.0pt]
        \bottomrule[1.0pt]
    \end{tabular}%
    }
    \caption{Experimental results on the \ourb~benchmark across different difficulty levels. We compare our \ourm~method with AlphaGeometry and the random baseline. In AlphaGeometry, we set the beam search parameter to be (beam, batch, depth) = (16, 64, 8). For both the random auxiliary point construction and the strategies in \ourm, we set $K$ to 2,048 and 8,192 to ensure a comprehensive evaluation.}
    \label{tab:benchmark}
\end{table*}

\begin{table}[t]
  \centering
  \resizebox{0.5\columnwidth}{!}{%
  \begin{tabular}{lc}
    \toprule[1.0pt]
    \toprule[1.0pt]
    Method & IMO-30 \\
    \midrule
    DDAR            & 15 \\
    AlphaGeometry       & ~24$^*$ \\
    Random Auxiliary points + DDAR       & 25 \\
    HAGeo   & 28 \\
    \bottomrule[1.0pt]
    \bottomrule[1.0pt]
  \end{tabular}
  }
  \vspace{-5pt}
  \caption{Comparison on the IMO-30 benchmark. $^*$ indicates that we report AlphaGeometry as solving 24 problems instead of 25, as we found the proof for IMO-2020-P1 to be incorrect, details provided in Appendix~\ref{sec:AlphaGeometry-false-positive}.}
  \label{tab:imo30}
\end{table}

\subsection{Settings}
\label{sec:settings}
\paragraph{Parameters}.
In our proposed \ourm, we set the number of auxiliary construction rounds to $N = 6$ in each auxiliary attempt. 
In experiments on the IMO-30 dataset in Table~\ref{tab:imo30}, we use $K = 4096$ auxiliary construction attempts, which is significantly fewer than the equivalent auxiliary and DDAR attempts used in AlphaGeometry.
For comparison on \ourb, we set the number of attempts $K = 2048$ for both our heuristic methods and a random baseline for adding auxiliary points, as this yields a roughly comparable number of DDAR samples to those in the AlphaGeometry experiment. 
We also report the performance of our method and the random baseline at $K = 8192$ to evaluate the upper bound of our approach.

\paragraph{Device and Time Constraints}
All experiments were conducted on a 64-core CPU machine, with an additional 80 GB A100 GPU used for AlphaGeometry’s language model (LM).
Each DDAR run was given a 60-second time limit per problem, with a total of 1.5 hours allocated for experiments on the IMO-30 benchmark.
We implemented a parallel version of AlphaGeometry's DDAR code and separated its LM and DDAR inference. 
The LM was first executed for full beam search under a 1.5-hour time limit, generating a complete list of auxiliary proposals.
For a fair comparison, we then run their DDAR to solve the problems under another 1.5-hour time constraint. 
To perform all methods on a 64-core machine under a 1.5 hour time limit, we set AlphaGeometry's beam search parameter to $(\text{beam}, \text{batch}, \text{depth}) = (16, 64, 8)$, which approximately reaches the maximum time limit setting on the 64-core CPU device.

\subsection{Results}
\label{sec:results}

\begin{figure*}[t]
    \centering
    \includegraphics[width=\linewidth]{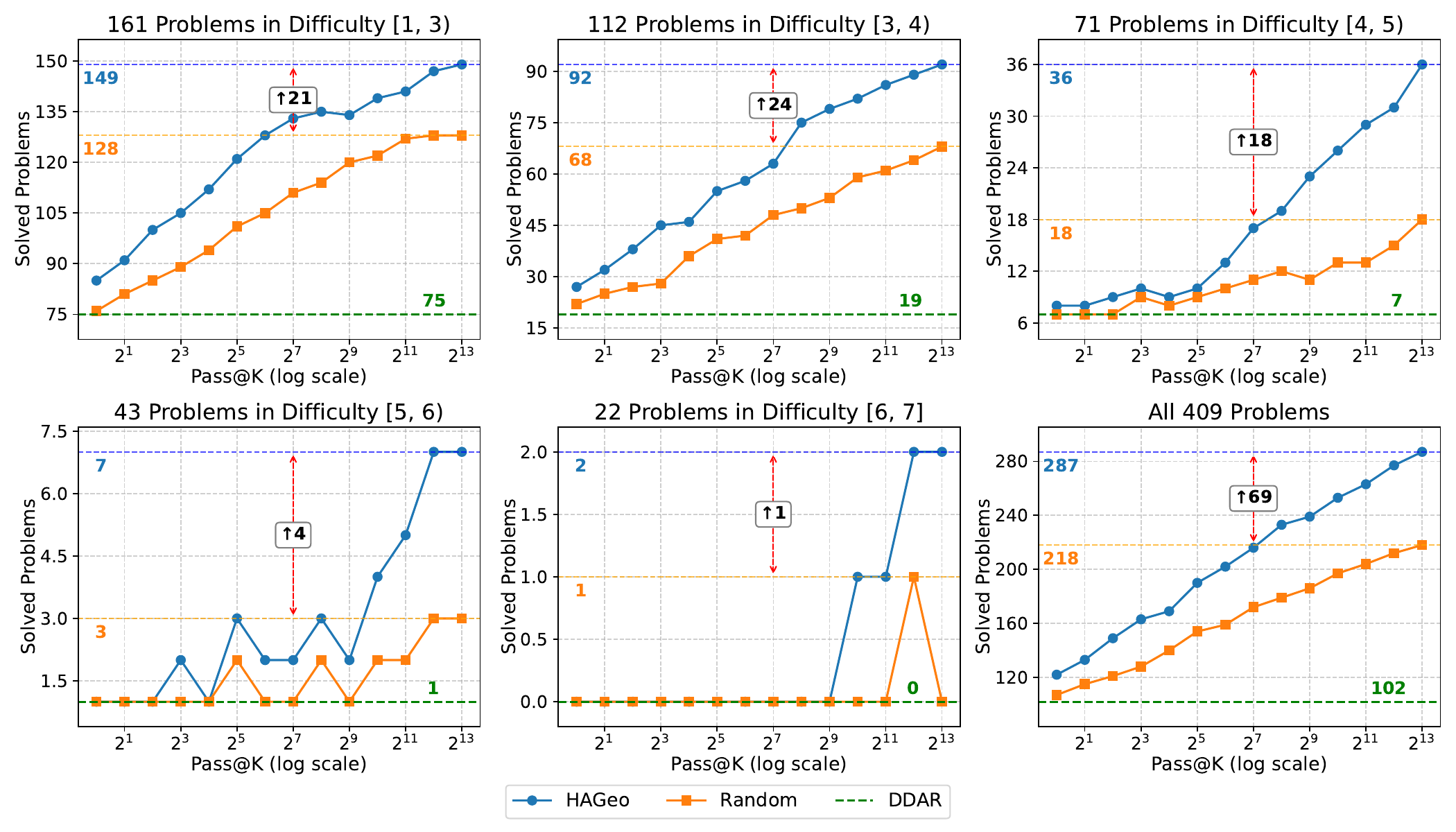}
    \caption{Pass@$K$ results of \ourm~and the random DDAR baseline on \ourb~across different difficulty levels. Top row: $[1,3)$, $[3,4)$, and $[4,5)$; Bottom row: $[5,6)$, $[6,7]$, and the full benchmark. The red double arrow \textcolor{red}{$\leftrightarrow$} indicates the performance gap between the best results of the random baseline and our method.}
    \label{fig:ablation}
\end{figure*}

\paragraph{Evaluation on the IMO-30 Benchmark}
We compare the performance of \ourm~with AlphaGeometry and the random baseline on the IMO-30 benchmark, as shown in Table~\ref{tab:imo30}.
Given the erroneous proofs potentially resulting from implementation mistakes in AlphaGeometry, we emphasize that manual verification is essential for the final evaluation. 
Therefore, we manually verified the proofs produced by our method on the IMO-30 benchmark, conducted by an IMO gold medalist. 
We find that even the random strategy for adding auxiliary points could solve 25 problems in the IMO-30 dataset, which is comparable to AlphaGeometry and reaches the ``silver-medal'' performance. 
Notably, our proposed \ourm~solves 28 out of 30 problems, outperforming AlphaGeometry by a notable margin and achieving ``gold-medal'' performance on IMO-30.
We further evaluated our method on recent IMO problems from 2024 and 2025 (with difficulty scores of 2.5 and 3.2, respectively), where it successfully solved them within 20 seconds each.

\paragraph{Evaluation on our \ourb~benchmark}.
We also compare our method with AlphaGeometry and the random baseline on the \ourb~benchmark, with the results reported in Table~\ref{tab:benchmark}.
Notably, all methods solve fewer problems as the difficulty increases, demonstrating the reliability of the difficulty annotations in \ourb.
Notably, our method performs effectively in the difficulty ranges [1,3) and [3,4) when $K$ is set to 8192, achieving solve rates of \textit{92.5\% and 83.0\%}, respectively, surpassing AlphaGeometry by \textit{19.2\% and 43.7\%}. 
Moreover, our method is capable of successfully solving two of the most challenging problems at the difficulty level $[6,7]$.
These results highlight both the robust generalization and the strong upper-bound performance of \ourm~in automated geometric theorem proving.

\subsection{Ablation Study}
We conduct an ablation study to analyze the effect of the number of attempts $K$ (Pass@$K$) on the \ourb~benchmark across different difficulty levels. 
Specifically, we report the number of solved problems under $K$ generated auxiliary constructions and DDAR attempts using both the random strategy and the strategy employed in \ourm, as illustrated in Figure~\ref{fig:ablation}.
As $K$ increases, both the random and our heuristic auxiliary point construction strategies show improved performance, with our approach consistently outperforming the random baseline. 
Moreover, our strategy is able to solve the most difficult problems in the level of $[6,7]$ when $K$ is sufficiently large. 
These results highlight the robust scalability and effectiveness of our strategy in addressing the most challenging geometry theorem-proving tasks as $K$ grows.

\section{Conclusion}
In this work, we propose \ourm, a heuristic method based on numerical computation using only CPUs to add auxiliary points in DDAR for automated Euclidean geometry theorem proving. 
Our method solves 28 out of 30 problems on the IMO-30 benchmark, achieving ``gold-medal'' performance while being 20 times faster than AlphaGeometry.
Furthermore, we develop a more comprehensive benchmark, \ourb, with human-evaluated difficulty annotations, featuring geometric problems that are generally more difficult than those in the IMO-30 benchmark.
Experiments on this new benchmark also demonstrate both the scalability and effectiveness of our method.
\clearpage
\bibliographystyle{assets/plainnat}
\bibliography{reference}

\begin{thebibliography}{26}
\providecommand{\natexlab}[1]{#1}
\providecommand{\url}[1]{\texttt{#1}}
\expandafter\ifx\csname urlstyle\endcsname\relax
  \providecommand{\doi}[1]{doi: #1}\else
  \providecommand{\doi}{doi: \begingroup \urlstyle{rm}\Url}\fi

\bibitem[Bak et~al.(2020)Bak, Kraj{\v{c}}i, and Rol{\'\i}nek]{bak2020automated}
P~Bak, RNDr~Stanislav Kraj{\v{c}}i, and Mgr~Michal Rol{\'\i}nek.
\newblock \emph{Automated generation of planar geometry Olympiad problems}.
\newblock PhD thesis, Master Thesis, 2020.

\bibitem[Chervonyi et~al.(2025)Chervonyi, Trinh, Ol{\v{s}}{\'a}k, Yang, Nguyen, Menegali, Jung, Verma, Le, and Luong]{chervonyi2025gold}
Yuri Chervonyi, Trieu~H Trinh, Miroslav Ol{\v{s}}{\'a}k, Xiaomeng Yang, Hoang Nguyen, Marcelo Menegali, Junehyuk Jung, Vikas Verma, Quoc~V Le, and Thang Luong.
\newblock Gold-medalist performance in solving olympiad geometry with alphageometry2.
\newblock \emph{arXiv preprint arXiv:2502.03544}, 2025.

\bibitem[Chou(1988)]{chou1988introduction}
Shang-Ching Chou.
\newblock An introduction to wu's method for mechanical theorem proving in geometry.
\newblock \emph{Journal of Automated Reasoning}, 4\penalty0 (3):\penalty0 237--267, 1988.

\bibitem[Chou et~al.(1993)Chou, Gao, and Zhang]{chou1993automated}
Shang-Ching Chou, Xiao-Shan Gao, and Jing-Zhong Zhang.
\newblock Automated production of traditional proofs for theorems in euclidean geometry.
\newblock In \emph{Proceedings of Eigth IEEE Symposium on Login in Computer Science}, pages 48--56, 1993.

\bibitem[Chou et~al.(2000)Chou, Gao, and Zhang]{chou2000deductive}
Shang-Ching Chou, Xiao-Shan Gao, and Jing-Zhong Zhang.
\newblock A deductive database approach to automated geometry theorem proving and discovering.
\newblock \emph{Journal of Automated Reasoning}, 25\penalty0 (3):\penalty0 219--246, 2000.

\bibitem[De~Moura et~al.(2015)De~Moura, Kong, Avigad, Van~Doorn, and von Raumer]{de2015lean}
Leonardo De~Moura, Soonho Kong, Jeremy Avigad, Floris Van~Doorn, and Jakob von Raumer.
\newblock The lean theorem prover (system description).
\newblock In \emph{Automated Deduction-CADE-25: 25th International Conference on Automated Deduction, Berlin, Germany, August 1-7, 2015, Proceedings 25}, pages 378--388. Springer, 2015.

\bibitem[Gelernter(1959)]{gelernter1959realization}
Herbert Gelernter.
\newblock Realization of a geometry-theorem proving machine.
\newblock In \emph{Proceedings of the International Conference on Information Processing}, page 273–282, 1959.

\bibitem[Gelernter et~al.(1960)Gelernter, Hansen, and Loveland]{gelernter1960empirical}
Herbert Gelernter, James~R Hansen, and Donald~W Loveland.
\newblock Empirical explorations of the geometry theorem machine.
\newblock In \emph{Papers presented at the May 3-5, 1960, western joint IRE-AIEE-ACM computer conference}, pages 143--149, 1960.

\bibitem[Hurst et~al.(2024)Hurst, Lerer, Goucher, Perelman, Ramesh, Clark, Ostrow, Welihinda, Hayes, Radford, et~al.]{hurst2024gpt}
Aaron Hurst, Adam Lerer, Adam~P Goucher, Adam Perelman, Aditya Ramesh, Aidan Clark, AJ~Ostrow, Akila Welihinda, Alan Hayes, Alec Radford, et~al.
\newblock Gpt-4o system card.
\newblock \emph{arXiv preprint arXiv:2410.21276}, 2024.

\bibitem[Lazard(1983)]{lazard1983grobner}
Daniel Lazard.
\newblock Gr{\"o}bner bases, gaussian elimination and resolution of systems of algebraic equations.
\newblock In \emph{European Conference on Computer Algebra}, pages 146--156. Springer, 1983.

\bibitem[Li et~al.(2025)Li, Zhang, Zhang, Zhang, Liu, Yao, Xu, Zheng, Wang, Chen, et~al.]{li2025system}
Zhong-Zhi Li, Duzhen Zhang, Ming-Liang Zhang, Jiaxin Zhang, Zengyan Liu, Yuxuan Yao, Haotian Xu, Junhao Zheng, Pei-Jie Wang, Xiuyi Chen, et~al.
\newblock From system 1 to system 2: A survey of reasoning large language models.
\newblock \emph{arXiv preprint arXiv:2502.17419}, 2025.

\bibitem[Liang et~al.(2025{\natexlab{a}})Liang, Li, Gong, Wang, Zhang, Shen, Wu, and Chen]{liang2025sws}
Xiao Liang, Zhong-Zhi Li, Yeyun Gong, Yang Wang, Hengyuan Zhang, Yelong Shen, Ying~Nian Wu, and Weizhu Chen.
\newblock Sws: Self-aware weakness-driven problem synthesis in reinforcement learning for llm reasoning.
\newblock \emph{arXiv preprint arXiv:2506.08989}, 2025{\natexlab{a}}.

\bibitem[Liang et~al.(2025{\natexlab{b}})Liang, Li, Gong, Shen, Wu, Guo, and Chen]{liang2025beyond}
Xiao Liang, Zhongzhi Li, Yeyun Gong, Yelong Shen, Ying~Nian Wu, Zhijiang Guo, and Weizhu Chen.
\newblock Beyond pass@ 1: Self-play with variational problem synthesis sustains rlvr.
\newblock \emph{arXiv preprint arXiv:2508.14029}, 2025{\natexlab{b}}.

\bibitem[Moura and Ullrich(2021)]{moura2021lean}
Leonardo~de Moura and Sebastian Ullrich.
\newblock The lean 4 theorem prover and programming language.
\newblock In \emph{International Conference on Automated Deduction}, pages 625--635. Springer, 2021.

\bibitem[Programs(2025)]{aops}
Aops Programs.
\newblock Art of problem solving.
\newblock \url{https://artofproblemsolving.com/community/c13_contests}, 2025.
\newblock Accessed: 2025-06-24.

\bibitem[Reiter(1972)]{reiter1972use}
Raymond Reiter.
\newblock \emph{The use of models in automatic theorem-proving}.
\newblock University of British Columbia. Department of Computer Science, 1972.

\bibitem[Sinha et~al.(2024)Sinha, Prabhu, Kumaraguru, Bhat, and Bethge]{sinha2024wu}
Shiven Sinha, Ameya Prabhu, Ponnurangam Kumaraguru, Siddharth Bhat, and Matthias Bethge.
\newblock Wu's method can boost symbolic ai to rival silver medalists and alphageometry to outperform gold medalists at imo geometry.
\newblock \emph{arXiv preprint arXiv:2404.06405}, 2024.

\bibitem[Sturm and Zengler(2011)]{JGEX231}
Thomas Sturm and Christoph Zengler.
\newblock Automated deduction in geometry: 7th international workshop, adg 2008, shanghai, china, september 22-24, 2008, revised papers, 04 2011.

\bibitem[Team(2025)]{geogebra}
Geogebra Team.
\newblock Geogebra.
\newblock \url{https://www.geogebra.org/}, 2025.
\newblock Accessed: 2025-06-24.

\bibitem[Trinh et~al.(2024)Trinh, Wu, Le, He, and Luong]{trinh2024solving}
Trieu~H Trinh, Yuhuai Wu, Quoc~V Le, He~He, and Thang Luong.
\newblock Solving olympiad geometry without human demonstrations.
\newblock \emph{Nature}, 625\penalty0 (7995):\penalty0 476--482, 2024.

\bibitem[Wei et~al.(2024)Wei, Sun, and Wang]{wei2024proving}
Chenrui Wei, Mengzhou Sun, and Wei Wang.
\newblock Proving olympiad algebraic inequalities without human demonstrations.
\newblock \emph{arXiv preprint arXiv:2406.14219}, 2024.

\bibitem[Wu(1978)]{wu1978decision}
W-T Wu.
\newblock On the decision problem and the mechanization of theorem proving in elementary geometry.
\newblock \emph{Scientia Sinica}, 21:\penalty0 157--179, 1978.

\bibitem[Xin et~al.(2024)Xin, Guo, Shao, Ren, Zhu, Liu, Ruan, Li, and Liang]{xin2024deepseek}
Huajian Xin, Daya Guo, Zhihong Shao, Zhizhou Ren, Qihao Zhu, Bo~Liu, Chong Ruan, Wenda Li, and Xiaodan Liang.
\newblock Deepseek-prover: Advancing theorem proving in llms through large-scale synthetic data.
\newblock \emph{arXiv preprint arXiv:2405.14333}, 2024.

\bibitem[Ying et~al.(2024)Ying, Wu, Geng, Wang, Lin, and Chen]{ying2024lean}
Huaiyuan Ying, Zijian Wu, Yihan Geng, Jiayu Wang, Dahua Lin, and Kai Chen.
\newblock Lean workbook: A large-scale lean problem set formalized from natural language math problems.
\newblock \emph{arXiv preprint arXiv:2406.03847}, 2024.

\bibitem[Zhang et~al.(2024)Zhang, Song, Li, Liang, Ma, Wang, Zhu, and Zhu]{zhang2024proposing}
Chi Zhang, Jiajun Song, Siyu Li, Yitao Liang, Yuxi Ma, Wei Wang, Yixin Zhu, and Song-Chun Zhu.
\newblock Proposing and solving olympiad geometry with guided tree search.
\newblock \emph{arXiv preprint arXiv:2412.10673}, 2024.

\bibitem[Zheng et~al.(2021)Zheng, Han, and Polu]{zheng2021minif2f}
Kunhao Zheng, Jesse~Michael Han, and Stanislas Polu.
\newblock Minif2f: a cross-system benchmark for formal olympiad-level mathematics.
\newblock \emph{arXiv preprint arXiv:2109.00110}, 2021.

\end{thebibliography}
\clearpage
\appendix
\label{sec:appendix}

\addtocontents{toc}{\protect\setcounter{tocdepth}{3}}
\renewcommand{\contentsname}{Appendix Contents for \ourm}
\hypersetup{linkcolor=black}
\tableofcontents 
\hypersetup{linkcolor=red}
\clearpage

\section{Fallacious Proof in AlphaGeometry}
\label{sec:AlphaGeometry-false-positive}
Here we present a detailed illustration of the fallacious proof for IMO-2020-P1 produced by AlphaGeometry. 
AlphaGeometry releases their proof to the IMO-30 problems in their supplementary material. As shown in section 2.28, the step 9 of the proof for IMO-2020-P1 is wrong.

\noindent In Step 9 of the proof, it states:\\

\begin{figure}[h]\footnotesize
\vspace{-15pt}
\begin{tcolorbox}[colback=gray!5!white, colframe=gray!60!black, boxsep=2pt, left=2pt, right=2pt, top=1.5pt, bottom=1.5pt]
$P, A, M$ are collinear, $A, Z, D$ are collinear, $A, D, E$ are collinear,
$\angle OPM = \angle ONM$ and $\angle ODE = \angle ONE \Rightarrow \angle (PO, AM) = \angle (DO, AZ)$.
\end{tcolorbox}
\vspace{-15pt}
\end{figure}

\noindent This statement is incorrect because it implicitly assumes that $M$, $N$, and $E$ are collinear, which is not proven.

\section{Preliminary for DDAR}
\label{sec:ddar-preliminary}
The Deductive Database and Algebraic Reasoning (\textbf{DDAR}) engine serves as the symbolic core of the existing generative automated geometry theorem-proving frameworks~\citep{trinh2024solving,zhang2024proposing,chervonyi2025gold}.
It combines rule-based deductive database (DD) inference with algebraic reasoning (AR), forming a unified system that jointly explores geometric and numerical dependencies.

\noindent \textbf{Deductive Database Engine}.
The deductive component focuses on inferring new geometric properties from known premises using a set of geometric rules that follow the paradigm ``If X, then Y''.
Each rule encodes a logically valid transformation, such as deducing collinearity, parallelism, or equality of segments and angles. 
Inspired by AlphaGeometry~\citep{trinh2024solving}, 
we adopt a structured deductive engine~\citep{chou2000deductive,JGEX231} that incrementally expands the deductive closure of all provable statements. 
This procedure is highly efficient, typically finishing in seconds on standard CPUs and requiring at most a few minutes, while ensuring comprehensive coverage of symbolic inferences.
A specific example of our deduction process in the geometry-specific language is provided in Case~\ref{case:examplary-output}.

\noindent \textbf{Algebraic Reasoning}.
The algebraic reasoning module complements the DD inference by handling arithmetic and ratio-based relations that cannot be captured symbolically. 
In AR, all geometric equalities are represented in a canonical linear form, such as $a - b - c + d = 0$, enabling the use of Gaussian elimination to uncover hidden dependencies among variables.
Variables may correspond to geometric quantities, including angles, distances, or ratios. 
As an example from AlphaGeometry, the angle equality $\angle ABC = \angle XYZ$ can be represented as $s(AB) - s(BC) = s(XY) - s(YZ)$, where $s(AB)$ denotes the angle between segment $AB$ and the x-axis, modulo $\pi$.
Notably, the algebraic computations can be represented in a coefficient matrix $A \in \mathbb{R}^{M \times N}$, where $N$ denotes the number of variables and $M$ the number of input equations.
To ensure numerical completeness, constant terms such as $\pi$, $\frac{1}{2}$, and rational ratios are also included in the coefficient matrix. 
This algebraic procedure efficiently generates new equivalences, such as quantitative relationships among existing angles or edges, thereby enhancing the symbolic reasoning process.

\noindent \textbf{Joint Deductive–Algebraic Inference}.
The DDAR engine could alternate between deductive and algebraic reasoning to build a consistent and exhaustive set of provable statements. 
After each round of symbolic deduction, the resulting statements could be passed to the algebraic module for further inference, and the new algebraic results are reintroduced into the deductive reasoning stage. 
This iterative interaction allows DDAR to reason across both geometric and numerical domains while maintaining logical soundness. 
In practice, the DDAR process converges rapidly, typically within a few seconds and may take up to minutes, yielding a compact yet complete closure of geometric knowledge for theorem proving.
\clearpage

\section{Example of our Geometry-Specific Language}
\label{sec:example}
Here we provide an example (IMO 2010 p2) illustrating how we convert a geometric problem from its natural language description into our geometry-specific language.

\noindent \textit{Problem in natural language:}
\begin{figure}[h]\footnotesize
\vspace{-5pt}
\begin{tcolorbox}[colback=gray!5!white, colframe=gray!60!black, boxsep=2pt, left=2pt, right=2pt, top=1.5pt, bottom=1.5pt]
Given a triangle $ABC$, with $I$ as its incenter and $\Gamma$ as its circumcircle, $AI$ intersects $\Gamma$ again at $D$. Let $E$ be a point on the arc $BDC$, and $F$ a point on the segment $BC$, such that $\angle BAF=\angle CAE < \dfrac12\angle BAC$. If $G$ is the midpoint of $IF$, prove that the meeting point of the lines $EI$ and $DG$ lies on $\Gamma$.
\end{tcolorbox}
\vspace{-15pt}
\end{figure}

\noindent \textit{Problem in our geometry-specific language:}
\begin{figure}[h]\footnotesize
\vspace{-5pt}
\begin{tcolorbox}[colback=msblue!5!white,colframe=msblue!80!black, boxsep=2pt, left=2pt, right=2pt, top=1.5pt, bottom=1.5pt]
\texttt{A B C = triangle; O = circumcenter A B C; (O) = circumcircle A B C; I = incenter A B C; AI = line A I; D = intersection AI (O); BC = line B C; F = on\_line BC; l = angle\_equal1 A C B A F; E = intersection l (O); G = midpoint I F; DG = line D G; EI = line E I; K = intersection DG EI; Prove: cong O A O K}
\end{tcolorbox}
\vspace{-15pt}
\end{figure}

\section{Example in our \ourb}
\noindent We also provide an example (ShuZhiMi 2023 MOST Mock p2) in our \ourb~benchmark.

\noindent \textit{Problem in natural language:}
\begin{figure}[htb]\footnotesize
\vspace{-5pt}
\begin{tcolorbox}[colback=gray!5!white, colframe=gray!60!black, boxsep=2pt, left=2pt, right=2pt, top=1.5pt, bottom=1.5pt]
Given two positively similar triangles $\triangle A_1 A_3 A_5 \sim \triangle A_4 A_6 A_2$. For $1 \le i \le 6$, define $X_i$ as the intersection of $A_iA_{i+2}$ and $A_{i+1}A_{i-1}$. For $1 \le i \le 6$, let $O_i$ be the circumcenter of $\triangle A_i X_i A_{i+1}$ ($A_7 = A_1$). Prove that the three lines $O_1 O_4, O_2 O_5, O_3 O_6$ are concurrent.
\end{tcolorbox}
\vspace{-15pt}
\end{figure}

\noindent \textit{Problem in our geometry-specific language:}

\begin{figure}[H]\footnotesize
\vspace{-5pt}
\begin{tcolorbox}[colback=msblue!5!white,colframe=msblue!80!black, boxsep=2pt, left=2pt, right=2pt, top=1.5pt, bottom=1.5pt]
\texttt{A1 A3 A5 = triangle; A4 = point; A6 = point; l1 = angle\_equal1 A4 A6 A5 A1 A3; l2 = angle\_equal1 A6 A4 A5 A3 A1; A2 = intersection l1 l2; A1A3 = line A1 A3; A2A6 = line A2 A6; X1 = intersection A1A3 A2A6; A2A4 = line A2 A4; X2 = intersection A1A3 A2A4; A3A5 = line A3 A5; X3 = intersection A2A4 A3A5; A4A6 = line A4 A6; X4 = intersection A3A5 A4A6; A1A5 = line A1 A5; X5 = intersection A1A5 A4A6; X6 = intersection A1A5 A2A6; O1 = circumcenter A1 X1 A2; O2 = circumcenter A2 X2 A3; O3 = circumcenter A3 X3 A4; O4 = circumcenter A4 X4 A5; O5 = circumcenter A5 X5 A6; O6 = circumcenter A6 X6 A1; O1O4 = line O1 O4; O2O5 = line O2 O5; K = intersection O1O4 O2O5; Prove: collinear K O3 O6}
\end{tcolorbox}
\end{figure}

\lstset{
  basicstyle=\ttfamily\scriptsize,
  breaklines=true,
  postbreak={},
  breakindent=0pt,
  frame=single,
  backgroundcolor=\color{gray!5},
  showstringspaces=false,
  keepspaces=false,
  columns=fullflexible,
  xleftmargin=0pt,
  framexleftmargin=0pt,
  literate={ω}{{$\omega$}}1 {Ω}{{$\Omega$}}1 {⊥}{{$\perp$}}1 {≡}{{$\equiv$}}1 {∠}{{$\angle$}}1 {Γ}{{$\Gamma$}}1 {γ}{{$\gamma$}}1,  
}

\clearpage
\onecolumn
\section{Example Proof of a Problem in IMO-30}
\begin{lstlisting}[caption={Example geometric theorem proving from our \ourm.},
label={case:examplary-output},captionpos=b,basicstyle=\ttfamily\scriptsize\linespread{0.9}\selectfont]

Problem IMO-2008-p6 in Natual Language (Converted following AlphaGeometry):
Let T1T2T3 be a triangle. Define point I as the circumcenter of triangle
T1T2T3. Define circle (I) as the circle centered I passing through T1. Let T4 be any point on circle (I). Define lines l1,l2,l3,l4 as the tangent lines at T1,T2,T3,T4 on circle (I). Define point A as the intersection of l1, l2. Define point B as the intersection of l2,l3. Define point C as the intersection of l3,l4. Define point D as the intersection of l4,l1. 
Define point I1 as incenter of triangle ABD. Define point I2 as the incenter of triangle BCD. Define point P1 as the foot of I1 on line BD. Define point P2 as the foot of I2 on line BD. Define (I1) as the circle centered I1 passing through P1. Define (I2) as the circle centered I2 passing through P2. Define lines m1, m2 be the excommon tangent of circle (I1), (I2). Define U1,U2 as the tangent points of m1, m2 with (I1). Define V1,V2 as the tangent points of m1, m2 with (I2). Define point X as the intersection of lines m1, m2. Prove that IX = IT1.

The original Problem:
T1 T2 T3 = obtuse_triangle; I = circumcenter T1 T2 T3; (I) = circle_center_point I T1; T4 = on_circle (I); IT1 = line I T1; IT2 = line I T2; IT3 = line I T3; IT4 = line I T4; l1 = perpendicular_line T1 IT1; l2 = perpendicular_line T2 IT2; A = intersection l1 l2; l3 = perpendicular_line T3 IT3; B = intersection l2 l3; l4 = perpendicular_line T4 IT4; C = intersection l3 l4; D = intersection l1 l4; I1 = incenter A B D; I2 = incenter B C D; BD = line B D; P1 = foot I1 BD; P2 = foot I2 BD; (I1) = circle_center_point I1 P1; (I2) = circle_center_point I2 P2; m1 m2 = ex_common_tangent (I1) (I2); U1 = intersection m1 (I1); U2 = intersection m2 (I1); V1 = intersection m1 (I2); V2 = intersection m2 (I2); X = intersection m1 m2; Prove: cong I X I T1

The problem with our Auxiliary constructions:
T1 T2 T3 = obtuse_triangle; I = circumcenter T1 T2 T3; (I) = circle_center_point I T1; T4 = on_circle (I); IT1 = line I T1; IT2 = line I T2; IT3 = line I T3; IT4 = line I T4; l1 = perpendicular_line T1 IT1; l2 = perpendicular_line T2 IT2; A = intersection l1 l2; l3 = perpendicular_line T3 IT3; B = intersection l2 l3; l4 = perpendicular_line T4 IT4; C = intersection l3 l4; D = intersection l1 l4; I1 = incenter A B D; I2 = incenter B C D; BD = line B D; P1 = foot I1 BD; P2 = foot I2 BD; (I1) = circle_center_point I1 P1; (I2) = circle_center_point I2 P2; m1 m2 = ex_common_tangent (I1) (I2); U1 = intersection m1 (I1); U2 = intersection m2 (I1); V1 = intersection m1 (I2); V2 = intersection m2 (I2); X = intersection m1 m2; l5 = line B I2; E = foot I l5; l6 = line I C; F = foot B l6; Prove: cong I X I T1

Proof:
IT1=IT2, IT1=IT3, IT1=IT4 => T4 on circle (T1T2T3)
<EI,l5>=<l6,BF>, F on line l6, E on line l5, I on line l6, B on line l5 => I on circle (BEF)
<l3,IT3>=<l4,IT4>, T3 on line l3, T4 on line l4, C on line l3, C on line l4 => T4 on circle (CIT3)
I is circumcenter of (T1T2T3), T4 on circle (T1T2T3), T4 on line l4, IT4⊥l4 => l4 is tangent to (T1T2T3)
I is circumcenter of (T1T2T3), T3 on line l3, IT3⊥l3 => l3 is tangent to (T1T2T3)
<IT3,l3>=<l6,BF>, F on line l6, T3 on line l3, I on line l6, B on line l3, I on circle (BEF) => T3 on circle (BEF)
<IT2,l2>=<l6,BF>, F on line l6, T2 on line l2, I on line l6, B on line l2, I on circle (BEF) => T2 on circle (BEF)
I is circumcenter of (T1T2T3), T2 on line l2, IT2⊥l2 => l2 is tangent to (T1T2T3)
T4 on circle (CIT3), I on line l6, C on line l6, T3 on line l3, C on line l3, C on line l4, T4 on line l4 => <IT3,l6>=<T3T4,l4>
l4 is tangent to (T1T2T3), T4 on circle (T1T2T3), T4 on line l4 => <T1T3,T1T4>=<T3T4,l4>
l3 is tangent to (T1T2T3), T3 on line l3, T4 on circle (T1T2T3) => <T1T3,T1T4>=<l3,T3T4>
I on circle (BEF), T3 on circle (BEF), F on line l6, I on line l6, T3 on line l3, B on line l3 => <IT3,l6>=<l3,BF>
I on circle (BEF), T2 on circle (BEF), T3 on circle (BEF), F on line l6, I on line l6 => <IT3,l6>=<T2T3,FT2>
l2 is tangent to (T1T2T3), T2 on line l2 => <T1T2,T1T3>=<l2,T2T3>
T2 on circle (BEF), T3 on circle (BEF), T2 on line l2, B on line l2, T3 on line l3, B on line l3 => <BF,FT3>=<l2,T2T3>
l3 is tangent to (T1T2T3), T3 on line l3 => <T1T2,T1T3>=<T2T3,l3>
<IT3,l6>=<l3,BF>, <IT3,l6>=<T3T4,l4>, <T1T3,T1T4>=<T3T4,l4>, <T1T3,T1T4>=<l3,T3T4> => <l3,BF>=<l3,T3T4>
<IT3,l6>=<T2T3,FT2>, <IT3,l6>=<T3T4,l4> => <T2T3,FT2>=<T3T4,l4>
<BF,FT3>=<l2,T2T3>, <T1T2,T1T3>=<l2,T2T3>, <T1T2,T1T3>=<T2T3,l3> => <BF,FT3>=<T2T3,l3>
<l3,BF>=<l3,T3T4> => BF//T3T4
I on circle (BEF), T2 on circle (BEF), T3 on circle (BEF), F on line l6, I on line l6 => <FT2,l6>=<T2T3,IT3>
I on circle (BEF), T2 on circle (BEF), T3 on circle (BEF), F on line l6, I on line l6 => <IT2,T2T3>=<l6,FT3>
IT2=IT3 => <IT2,T2T3>=<T2T3,IT3>
<T2T3,FT2>=<T3T4,l4> => <T3T4,T2T3>=<l4,FT2>
<BF,FT3>=<T2T3,l3> => <BF,T2T3>=<FT3,l3>
BF//T3T4 => <BF,T2T3>=<T3T4,T2T3>
<FT2,l6>=<T2T3,IT3>, <IT2,T2T3>=<T2T3,IT3>, <IT2,T2T3>=<l6,FT3> => <FT2,l6>=<l6,FT3>
<BF,T2T3>=<FT3,l3>, <BF,T2T3>=<T3T4,T2T3>, <T3T4,T2T3>=<l4,FT2> => <FT3,l3>=<l4,FT2>
T3 on circle (BEF), E on line l5, B on line l5, T3 on line l3, B on line l3 => <BF,FT3>=<l5,ET3>
T2 on circle (BEF), T3 on circle (BEF) => <FT3,EF>=<T2T3,ET2>
T3 on circle (BEF), E on line l5, B on line l5, T3 on line l3, B on line l3 => <FT3,EF>=<l3,l5>
<FT2,l6>=<l6,FT3> => <FT2,l6>=<l6,FT3>
<CI2,l4>=<l3,CI2> => <CI2,l3>=<l4,CI2>
<FT3,l3>=<l4,FT2> => <FT2,l3>=<l4,FT3>
<T1T2,T1T3>=<T2T3,l3>, <T1T2,T1T3>=<l2,T2T3>, <BF,FT3>=<l2,T2T3>, <BF,FT3>=<l5,ET3> => <T2T3,l3>=<l5,ET3>
<FT3,EF>=<T2T3,ET2>, <FT3,EF>=<l3,l5>, <l3,l5>=<l5,BD> => <T2T3,ET2>=<l5,BD>
<CI2,l3>=<l4,CI2>, <FT2,l3>=<l4,FT3>, <FT2,l6>=<l6,FT3> => <CI2,l6>=<l6,CI2>
<T2T3,l3>=<l5,ET3> => <T2T3,l5>=<l3,ET3>
T2 on circle (BEF), T2 on line l2, B on line l2, E on line l5, B on line l5 => <BF,EF>=<l2,ET2>
T3 on circle (BEF), E on line l5, B on line l5, T3 on line l3, B on line l3 => <BF,EF>=<l3,ET3>
<T2T3,ET2>=<l5,BD> => <ET2,BD>=<T2T3,l5>
<CI2,l6>=<l6,CI2> => CI2//l6
<ET2,BD>=<T2T3,l5>, <T2T3,l5>=<l3,ET3>, <BF,EF>=<l3,ET3>, <BF,EF>=<l2,ET2> => <ET2,BD>=<l2,ET2>
CI2//l6, C on line l6 => line CI2≡l6
<BI1,BD>=<l2,BI1> => <BD,BI1>=<BI1,l2>
<ET2,BD>=<l2,ET2> => <BD,ET2>=<ET2,l2>
line CI2≡l6 => I2 on line l6
<BD,BI1>=<BI1,l2>, <BD,ET2>=<ET2,l2> => <BI1,ET2>=<ET2,BI1>
I on circle (BEF), T3 on circle (BEF), F on line l6, I on line l6 => <FT3,EF>=<IT3,EI>
I on circle (BEF), T3 on circle (BEF), F on line l6, I on line l6, T3 on line l3, B on line l3 => <BF,FT3>=<BI,IT3>
T2 on circle (BEF), T3 on circle (BEF), T2 on line l2, B on line l2, T3 on line l3, B on line l3 => <FT2,BF>=<T2T3,l3>
T2 on circle (BEF), T2 on line l2, B on line l2, E on line l5, B on line l5 => <ET2,l5>=<FT2,BF>
<BD,I2P2>=<BF,l6>, P2 on line BD, F on line l6, I2 on line l6 => P2 on circle (BFI2)
<BI1,ET2>=<ET2,BI1> => BI1//ET2
<FT3,EF>=<IT3,EI>, <FT3,EF>=<l3,l5>, <l3,l5>=<l5,BD> => <IT3,EI>=<l5,BD>
<BF,FT3>=<BI,IT3>, <BF,FT3>=<l2,T2T3>, <T1T2,T1T3>=<l2,T2T3>, <T1T2,T1T3>=<T2T3,l3>, <FT2,BF>=<T2T3,l3>, <ET2,l5>=<FT2,BF> => <BI,IT3>=<ET2,l5>
P2 on circle (BFI2), F on line l6, I2 on line l6, P2 on line BD, I2 on line l5, B on line l5 => <FP2,BF>=<I2P2,l5>
<BD,I2P2>=<EI,l5> => <BD,EI>=<I2P2,l5>
BI1//ET2 => <BI1,BI>=<ET2,BI>
<IT3,EI>=<l5,BD> => <BD,EI>=<l5,IT3>
<BI,IT3>=<ET2,l5> => <ET2,BI>=<l5,IT3>
l2 is tangent to (T1T2T3), T2 on line l2, T4 on circle (T1T2T3) => <T2T4,T3T4>=<l2,T2T3>
I on circle (BEF), T2 on circle (BEF), F on line l6, I on line l6, T2 on line l2, B on line l2 => <FT2,BF>=<IT2,BI>
<BI1,BI>=<ET2,BI>, <ET2,BI>=<l5,IT3>, <BD,EI>=<l5,IT3>, <BD,EI>=<I2P2,l5>, <FP2,BF>=<I2P2,l5> => <BI1,BI>=<FP2,BF>
<FT2,BF>=<IT2,BI>, <FT2,BF>=<T2T3,l3>, <T1T2,T1T3>=<T2T3,l3>, <T1T2,T1T3>=<l2,T2T3>, <T2T4,T3T4>=<l2,T2T3> => <IT2,BI>=<T2T4,T3T4>
<BI1,BI>=<FP2,BF> => <BF,BI>=<FP2,BI1>
<IT2,BI>=<T2T4,T3T4> => <T2T4,IT2>=<T3T4,BI>
BF//T3T4 => <BF,BI>=<T3T4,BI>
IT2=IT4 => <IT4,T2T4>=<T2T4,IT2>
<BF,BI>=<FP2,BI1>, <BF,BI>=<T3T4,BI>, <T2T4,IT2>=<T3T4,BI>, <IT4,T2T4>=<T2T4,IT2> => <FP2,BI1>=<IT4,T2T4>
<ET2,l5>=<FT2,BF>, <FT2,BF>=<T2T3,l3>, <T1T2,T1T3>=<T2T3,l3>, <T1T2,T1T3>=<l2,T2T3>, <T2T4,T3T4>=<l2,T2T3> => <ET2,l5>=<T2T4,T3T4>
<FP2,BI1>=<IT4,T2T4> => <IT4,FP2>=<T2T4,BI1>
BF//T3T4 => <BF,l5>=<T3T4,l5>
<ET2,l5>=<T2T4,T3T4> => <T2T4,ET2>=<T3T4,l5>
P2 on circle (BFI2), F on line l6, I2 on line l6, P2 on line BD, I2 on line l5, B on line l5 => <BF,l5>=<FP2,I2P2>
BI1//ET2 => <T2T4,BI1>=<T2T4,ET2>
<BF,l5>=<FP2,I2P2>, <BF,l5>=<T3T4,l5>, <T2T4,ET2>=<T3T4,l5>, <T2T4,BI1>=<T2T4,ET2>, <IT4,FP2>=<T2T4,BI1> => <FP2,I2P2>=<IT4,FP2>
<BD,I2P2>=<IT4,l4> => <BD,I2P2>=<IT4,l4>
<BD,DI2>=<DI2,l4> => <BD,DI2>=<DI2,l4>
<FP2,I2P2>=<IT4,FP2> => <FP2,I2P2>=<IT4,FP2>
m2 is tangent to (I1), I1 is circumcenter of (I1), U2 on circle (I1), U2 on line m2 => I1U2⊥m2
m1 is tangent to (I1), I1 is circumcenter of (I1), U1 on circle (I1), U1 on line m1 => I1U1⊥m1
m1 is tangent to (I2), I2 is circumcenter of (I2), V1 on circle (I2), V1 on line m1 => I2V1⊥m1
m2 is tangent to (I2), I2 is circumcenter of (I2), V2 on circle (I2), V2 on line m2 => I2V2⊥m2
<BD,DI2>=<DI2,l4>, <BD,I2P2>=<IT4,l4>, <FP2,I2P2>=<IT4,FP2> => <DI2,FP2>=<FP2,DI2>
<l1,IT1>=<l2,IT2>, T2 on line l2, T1 on line l1, A on line l2, A on line l1 => T2 on circle (AIT1)
<DI2,FP2>=<FP2,DI2> => DI2⊥FP2
<IT3,l6>=<T3T4,l4>, <T1T3,T1T4>=<T3T4,l4>, <T1T3,T1T4>=<l3,T3T4> => <IT3,l6>=<l3,T3T4>
T2 on circle (AIT1), T2 on line l2, A on line l2, A on line l1, T1 on line l1 => <T1T2,IT2>=<l1,AI>
T2 on circle (AIT1), T2 on line l2, A on line l2, A on line l1, T1 on line l1 => <AI,l2>=<IT1,T1T2>
IT1=IT2 => <IT1,T1T2>=<T1T2,IT2>
<l1,IT1>=<l4,IT4>, T1 on line l1, T4 on line l4, D on line l1, D on line l4 => T4 on circle (DIT1)
m1⊥I1U1, m2⊥I1U2 => <m1,I1U1>=<m2,I1U2>
I2V1⊥m1, I2V2⊥m2 => <I2V1,m1>=<I2V2,m2>
<IT3,l6>=<l3,T3T4> => <T3T4,l6>=<l3,IT3>
<AI,l2>=<IT1,T1T2>, <IT1,T1T2>=<T1T2,IT2>, <T1T2,IT2>=<l1,AI> => <AI,l2>=<l1,AI>
T4 on circle (DIT1), D on line l1, T1 on line l1, D on line l4, T4 on line l4 => <DI,IT1>=<l4,T1T4>
l4 is tangent to (T1T2T3), T4 on circle (T1T2T3), T4 on line l4 => <T2T4,T1T2>=<l4,T1T4>
<m1,I1U1>=<m2,I1U2> => <I1U1,I1U2>=<m1,m2>
<I2V1,m1>=<I2V2,m2> => <I2V1,I2V2>=<m1,m2>
m2⊥I1U2, m2⊥I2V2 => <m2,I1U2>=<m2,I2V2>
DI2⊥FP2, l3⊥IT3, <T3T4,l6>=<l3,IT3> => <DI2,FP2>=<T3T4,l6>
<AI1,l1>=<l2,AI1> => <AI1,l1>=<l2,AI1>
<AI,l2>=<l1,AI> => <AI,l1>=<l2,AI>
<DI,IT1>=<l4,T1T4>, <T2T4,T1T2>=<l4,T1T4> => <DI,IT1>=<T2T4,T1T2>
I1U2=I1U1, I2V1=I2V2 => I1U2/I1U1=I2V1/I2V2
<I1U1,I1U2>=<m1,m2>, <I2V1,I2V2>=<m1,m2> => <I1U1,I1U2>=<I2V1,I2V2>
<m2,I1U2>=<m2,I2V2> => I1U2//I2V2
FP2⊥DI2, l3⊥IT3, <T3T4,l6>=<l3,IT3> => <FP2,DI2>=<T3T4,l6>
<BF,l5>=<FP2,I2P2>, <BF,l5>=<T3T4,l5> => <FP2,I2P2>=<T3T4,l5>
P2 on circle (BFI2), F on line l6, I2 on line l6, P2 on line BD, I2 on line l5, B on line l5 => <BD,l5>=<FP2,l6>
<DI2,FP2>=<T3T4,l6> => <DI2,T3T4>=<FP2,l6>
T2 on circle (BEF), T3 on circle (BEF) => <EF,FT3>=<ET2,T2T3>
BI1//ET2 => <BI1,T2T3>=<ET2,T2T3>
T3 on circle (BEF), E on line l5, B on line l5, T3 on line l3, B on line l3 => <EF,FT3>=<l5,l3>
<AI,l1>=<l2,AI>, <AI1,l1>=<l2,AI1> => <AI,AI1>=<AI1,AI>
<DI,IT1>=<T2T4,T1T2> => <DI,T2T4>=<IT1,T1T2>
I1U2/I1U1=I2V1/I2V2, <I1U1,I1U2>=<I2V1,I2V2> => <I2V1,V1V2>=<U1U2,I1U2>
I1U2//I2V2 => <V1V2,I1U2>=<V1V2,I2V2>
I2V1=I2V2 => <I2V1,V1V2>=<V1V2,I2V2>
I1U1=I1U2 => <I1U1,U1U2>=<U1U2,I1U2>
I1U1⊥m1, m2⊥I1U2 => <I1U1,m1>=<m2,I1U2>
<FP2,DI2>=<T3T4,l6> => <DI2,l6>=<FP2,T3T4>
<FP2,I2P2>=<T3T4,l5> => <FP2,T3T4>=<I2P2,l5>
<I1P1,BD>=<l5,EI> => <BD,EI>=<I1P1,l5>
<BI1,T2T3>=<ET2,T2T3>, <EF,FT3>=<ET2,T2T3>, <EF,FT3>=<l5,l3>, <BD,l5>=<l5,l3>, <BD,l5>=<FP2,l6>, <DI2,T3T4>=<FP2,l6> => <BI1,T2T3>=<DI2,T3T4>
<AI,AI1>=<AI1,AI> => AI//AI1
<AI,l2>=<IT1,T1T2>, <DI,T2T4>=<IT1,T1T2> => <AI,l2>=<DI,T2T4>
<I2V1,m1>=<I2V2,m2>, V2 on line m2, V1 on line m1, X on line m2, X on line m1 => X on circle (I2V1V2)
<m1,I1U1>=<m2,I1U2>, U2 on line m2, U1 on line m1, X on line m2, X on line m1 => X on circle (I1U1U2)
m1 is tangent to (I2), V1 on circle (I2), V1 on line m1, V2 on circle (I2), P2 on circle (I2) => <P2V1,P2V2>=<m1,V1V2>
m2 is tangent to (I2), V2 on circle (I2), V2 on line m2, V1 on circle (I2), P2 on circle (I2) => <P2V1,P2V2>=<V1V2,m2>
<I1U1,U1U2>=<U1U2,I1U2>, <I2V1,V1V2>=<U1U2,I1U2>, <I2V1,V1V2>=<V1V2,I2V2>, <V1V2,I1U2>=<V1V2,I2V2> => <I1U1,U1U2>=<V1V2,I1U2>
<I1U1,m1>=<m2,I1U2> => <I1U1,m2>=<m1,I1U2>
<DI2,l6>=<FP2,T3T4>, <FP2,T3T4>=<I2P2,l5> => <DI2,l6>=<I2P2,l5>
<BI1,BI>=<ET2,BI>, <ET2,BI>=<l5,IT3>, <BD,EI>=<l5,IT3>, <BD,EI>=<I1P1,l5> => <BI1,BI>=<I1P1,l5>
<BI1,T2T3>=<DI2,T3T4> => <BI1,DI2>=<T2T3,T3T4>
BF//T3T4 => <T2T3,BF>=<T2T3,T3T4>
<BF,FT3>=<T2T3,l3> => <T2T3,BF>=<l3,FT3>
I on circle (BEF), T3 on circle (BEF), F on line l6, I on line l6, T3 on line l3, B on line l3 => <BI,l6>=<l3,FT3>
AI//AI1 => line AI≡AI1
l2 is tangent to (T1T2T3), T2 on line l2, T4 on circle (T1T2T3) => <T2T3,T3T4>=<l2,T2T4>
<AI,l2>=<DI,T2T4> => <AI,DI>=<l2,T2T4>
X on circle (I2V1V2), V1 on line m1, X on line m1, V2 on line m2, X on line m2 => <V1V2,I2V2>=<m1,I2X>
X on circle (I1U1U2), U2 on line m2, X on line m2, X on line m1, U1 on line m1 => <U1U2,I1U2>=<m1,I1X>
<P2V1,P2V2>=<V1V2,m2>, <P2V1,P2V2>=<m1,V1V2> => <V1V2,m2>=<m1,V1V2>
<I1U1,U1U2>=<V1V2,I1U2> => <I1U1,U1U2>=<V1V2,I1U2>
<I1U1,m2>=<m1,I1U2> => <I1U1,m1>=<m2,I1U2>
<DI2,l6>=<I2P2,l5>, <BD,EI>=<I2P2,l5>, P2 on line BD, I2 on line l5, E on line l5, I2 on line l6, I on line l6 => DP2/DI2=EI/II2
<BI1,BI>=<I1P1,l5>, <BD,EI>=<I1P1,l5>, P1 on line BD, B on line l5, E on line l5 => BE/BI=I1P1/BI1
<DI2,l6>=<I2P2,l5>, <BD,EI>=<I2P2,l5>, P2 on line BD, I2 on line l5, E on line l5, I2 on line l6, I on line l6 => DP2/I2P2=EI/EI2
<BI,l6>=<l3,FT3>, <T2T3,BF>=<l3,FT3>, <T2T3,BF>=<T2T3,T3T4>, <BI1,DI2>=<T2T3,T3T4> => <BI,l6>=<BI1,DI2>
line AI≡AI1 => I1 on line AI
<AI,DI>=<l2,T2T4>, <T2T3,T3T4>=<l2,T2T4>, <T2T3,BF>=<T2T3,T3T4>, <T2T3,BF>=<l3,FT3>, <BI,l6>=<l3,FT3> => <AI,DI>=<BI,l6>
I1U1⊥m1, I2V1⊥m1 => <I1U1,m1>=<I2V1,m1>
<U1U2,I1U2>=<m1,I1X>, <I2V1,V1V2>=<U1U2,I1U2>, <I2V1,V1V2>=<V1V2,I2V2>, <V1V2,I2V2>=<m1,I2X> => <m1,I1X>=<m1,I2X>
<V1V2,m2>=<m1,V1V2> => <V1V2,m1>=<m2,V1V2>
<I1U1,U1U2>=<V1V2,I1U2>, <I1U1,m1>=<m2,I1U2> => <U1U2,m1>=<m2,V1V2>
I2V1=I2V2, I2V2=I2V1 => I2V1/I2V2=I2V2/I2V1
DP2/DI2=EI/II2 => EI/DP2=II2/DI2
BE/BI=I1P1/BI1 => BE/I1P1=BI/BI1
I1P1=I1U2 => BE/I1P1=BE/I1U2
DP2/I2P2=EI/EI2 => EI/DP2=EI2/I2P2
<BI,l6>=<BI1,DI2>, <AI,DI>=<BI,l6>, I1 on line AI, I2 on line l6, I on line l6 => BI/BI1=II2/DI2
<I1U1,m1>=<I2V1,m1> => I1U1//I2V1
<m1,I1X>=<m1,I2X> => I1X//I2X
<V1V2,m1>=<m2,V1V2>, X on line m2, V2 on line m2, X on line m1, V1 on line m1 => V1X/V1V2=V2X/V1V2
<U1U2,m1>=<m2,V1V2>, X on line m2, U2 on line m2, X on line m1, U1 on line m1, V1 on line m1, V2 on line m2 => U1X/U1U2=V2X/V1V2
I2V1/I2V2=I2V2/I2V1 => V1V2/I2V1=V1V2/I2V2
I1U2/I1U1=I2V1/I2V2, <I1U1,I1U2>=<I2V1,I2V2> => U1U2/I1U1=V1V2/I2V2
BE/I1P1=BE/I1U2, BE/I1P1=BI/BI1, BI/BI1=II2/DI2, EI/DP2=II2/DI2, EI/DP2=EI2/I2P2 => BE/I1U2=EI2/I2P2
I1U1//I2V1, I1X//I2X, U1 on line m1, X on line m1, V1 on line m1 => I1X/U1X=I2X/V1X
I1P1=I1U2, I2V1=I2P2 => I1P1/I1U2=I2V1/I2P2
U1X/U1U2=V2X/V1V2, V1X/V1V2=V2X/V1V2 => U1X/U1U2=V1X/V1V2
I1U1=I1P1, I2P2=I2V1 => I1U1/I1P1=I2P2/I2V1
U1U2/I1U1=V1V2/I2V2, V1V2/I2V1=V1V2/I2V2 => U1U2/I1U1=V1V2/I2V1
I1U1=I1P1, I2P2=I2V2 => I1U1/I1P1=I2P2/I2V2
I1U1=I1P1, I2V1=I2V2 => I1U1/I1P1=I2V1/I2V2
I1X//I2X => line I1X≡I2X
BE/I1U2=EI2/I2P2 => BE/EI2=I1U2/I2P2
I1X/U1X=I2X/V1X => I1X/I2X=U1X/V1X
I1P1/I1U2=I2V1/I2P2 => I1P1/I2V1=I1U2/I2P2
U1X/U1U2=V1X/V1V2 => U1U2/V1V2=U1X/V1X
I1U1/I1P1=I2P2/I2V1 => I1P1/I2V1=I1U1/I2P2
U1U2/I1U1=V1V2/I2V1 => I1U1/I2V1=U1U2/V1V2
I1U1/I1P1=I2P2/I2V2 => I1P1/I2V2=I1U1/I2P2
I1U1/I1P1=I2V1/I2V2 => I1P1/I2V2=I1U1/I2V1
line I1X≡I2X => I2 on line I1X
BE/EI2=I1U2/I2P2, I1P1/I2V1=I1U2/I2P2, I1P1/I2V1=I1U1/I2P2, I1P1/I2V2=I1U1/I2P2, I1P1/I2V2=I1U1/I2V1, I1U1/I2V1=U1U2/V1V2, U1U2/V1V2=U1X/V1X, I1X/I2X=U1X/V1X => BE/EI2=I1X/I2X
BE/EI2=I1X/I2X, I2 on line I1X, E on line l5, B on line l5, I2 on line l5, X I1 I2 and E B I2 have same order => BE/BI2=I1X/I1I2
BE/BI2=I1X/I1I2 => I1I2/BI2=I1X/BE
BE/EI2=I1X/I2X => I1X/BE=I2X/EI2
I1I2/BI2=I1X/BE, I1X/BE=I2X/EI2 => I1I2/BI2=I2X/EI2
FP2⊥DI2, I1P1⊥BD => <FP2,DI2>=<I1P1,BD>
<FP2,BF>=<I2P2,l5>, <BD,EI>=<I2P2,l5>, <BD,EI>=<I1P1,l5> => <FP2,BF>=<I1P1,l5>
I1I2/BI2=I2X/EI2, I2 on line I1X, I2 on line l5, E on line l5, B on line l5 => <l5,BI1>=<l5,EX>
I on circle (BEF), T2 on circle (BEF), F on line l6, I on line l6, T2 on line l2, B on line l2 => <BF,FT2>=<BI,IT2>
T2 on circle (BEF), T2 on line l2, B on line l2, E on line l5, B on line l5 => <BF,FT2>=<l5,ET2>
BI1//ET2 => <l5,BI1>=<l5,ET2>
I on circle (BEF), T2 on circle (BEF), F on line l6, I on line l6 => <FT2,EF>=<IT2,EI>
T2 on circle (BEF), T2 on line l2, B on line l2, E on line l5, B on line l5 => <FT2,EF>=<l2,l5>
<FT3,EF>=<l3,l5>, <l3,l5>=<l5,BD> => <FT3,EF>=<l5,BD>
I on circle (BEF), F on line l6, I on line l6, E on line l5, B on line l5 => <BF,l5>=<l6,EI>
<FP2,DI2>=<I1P1,BD> => <DI2,BD>=<FP2,I1P1>
I on circle (BEF), T3 on circle (BEF), F on line l6, I on line l6 => <FT3,ET3>=<l6,EI>
<FP2,BF>=<I1P1,l5> => <BF,l5>=<FP2,I1P1>
<BF,FT2>=<BI,IT2>, <BF,FT2>=<l5,ET2>, <l5,BI1>=<l5,ET2>, <l5,BI1>=<l5,EX> => <BI,IT2>=<l5,EX>
<FT2,EF>=<IT2,EI>, <FT2,EF>=<l2,l5> => <IT2,EI>=<l2,l5>
<FT3,EF>=<l5,BD> => <EF,BD>=<FT3,l5>
<DI2,BD>=<FP2,I1P1>, <BF,l5>=<FP2,I1P1>, <BF,l5>=<l6,EI>, <FT3,ET3>=<l6,EI> => <DI2,BD>=<FT3,ET3>
I on circle (BEF), F on line l6, I on line l6, E on line l5, B on line l5 => <EF,l5>=<l6,BI>
<BI,IT2>=<l5,EX> => <BI,IT2>=<l5,EX>
<IT2,EI>=<l2,l5> => <EI,IT2>=<l5,l2>
<EF,BD>=<FT3,l5> => <BD,EF>=<l5,FT3>
<DI2,BD>=<FT3,ET3> => <BD,DI2>=<ET3,FT3>
<BI,l6>=<BI1,DI2>, <AI,DI>=<BI,l6>, I1 on line AI, I2 on line l6, I on line l6 => DI/DI2=II1/BI1
<EF,l5>=<l6,BI> => <BI,l5>=<l6,EF>
<BI,l6>=<BI1,DI2>, <AI,DI>=<BI,l6>, I1 on line AI, I2 on line l6, I on line l6 => BI/II1=II2/DI
<BI,IT2>=<l5,EX>, <EI,IT2>=<l5,l2> => <BI,EI>=<l2,EX>
<BD,DI2>=<ET3,FT3>, <BD,EF>=<l5,FT3> => <DI2,EF>=<l5,ET3>
DI/DI2=II1/BI1 => DI/II1=DI2/BI1
<BI,l5>=<l6,EF>, I2 on line l6, I on line l6, I2 on line l5, B on line l5, E on line l5, F on line l6 => EI2/EF=II2/BI
BI/II1=II2/DI => DI/II1=II2/BI
IT2=IT3 => <IT3,T2T3>=<T2T3,IT2>
I on circle (BEF), T2 on circle (BEF), T3 on circle (BEF), F on line l6, I on line l6 => <IT3,T2T3>=<l6,FT2>
I on circle (BEF), T3 on circle (BEF), F on line l6, I on line l6 => <ET3,EI>=<FT3,l6>
I on circle (BEF), T2 on circle (BEF), T3 on circle (BEF), F on line l6, I on line l6 => <FT3,l6>=<T2T3,IT2>
<BI,EI>=<l2,EX> => <BI,l2>=<EI,EX>
I on circle (BEF), T2 on circle (BEF), F on line l6, I on line l6, T2 on line l2, B on line l2 => <BI,l2>=<l6,FT2>
<DI2,EF>=<l5,ET3>, <BD,EF>=<l5,FT3>, I2 on line l5, B on line l5 => DI2/BI2=ET3/EF
I1I2/BI2=I2X/EI2, I2 on line I1X, I2 on line l5, E on line l5, B on line l5 => BI2/BI1=EI2/EX
DI/II1=DI2/BI1, DI/II1=II2/BI, EI2/EF=II2/BI => DI2/BI1=EI2/EF
<BI,l2>=<EI,EX>, <BI,l2>=<l6,FT2>, <IT3,T2T3>=<l6,FT2>, <IT3,T2T3>=<T2T3,IT2>, <FT3,l6>=<T2T3,IT2>, <ET3,EI>=<FT3,l6> => <EI,EX>=<ET3,EI>
BI2/BI1=EI2/EX, DI2/BI2=ET3/EF, DI2/BI1=EI2/EF => EI/ET3=EI/EX
EI/ET3=EI/EX, <EI,EX>=<ET3,EI> => IT3/ET3=IX/EX
BI2/BI1=EI2/EX, DI2/BI2=ET3/EF, DI2/BI1=EI2/EF, IT2=IT4, IT3=IT4 => IT2/EX=IT3/ET3
IT2/EX=IT3/ET3, IT3/ET3=IX/EX => IT2/EX=IX/EX
IT2/EX=IX/EX => IT2=IX
IT1=IT2, IT2=IX => IT1=IX
IT1=IX => Goal

\end{lstlisting}

We manually convert and improve the proof of HAGeometry to make it easier to understand.
\begin{figure}[h]\footnotesize
\vspace{-5pt}
\begin{tcolorbox}[
  enhanced,
  width=\linewidth+0.26cm,   
  enlarge left by=-0.13cm, 
  enlarge right by=-0.13cm,
  colback=white,
  colframe=black,
  boxrule=0.6pt,
  sharp corners,
  boxsep=2pt,
  left=2pt, right=2pt, top=1.5pt, bottom=1.5pt
]
Proof: \\

$DP_2 = \frac{BD + CD - BC}{2} = \frac{BD + DT_4 - BT_3}{2} = \frac{BD + DT_1 - BT_2}{2} = \frac{BD + AB-AD}{2} = BP_1$. \\

$\angle EI_2I = 90^{\circ} - \frac{\angle BDC}{2} = \angle P_2I_2D \Rightarrow \triangle EII_2 \sim \triangle P_2DI_2$. \\

$\angle EBI = 90^{\circ} - \frac{\angle ABD}{2} = \angle P_1I_1B \Rightarrow \triangle BEI \sim \triangle I_1P_1B$ \\

$\frac{EI_2}{I_2P_2 } = \frac{EI}{DP_2} = \frac{EI}{BP_1} = \frac{BE}{I_1P_1}$ \\

$\frac{EI_2}{BE} = \frac{I_2P_2}{P_1P_1} = \frac{XI_2}{XI_1}$ \\

$\Rightarrow EX \parallel I_1B$ \\

$\Rightarrow \frac{EI_2}{BI_2} = \frac{EX}{BI_1}$ \\

$\angle BI_1I = 90^{\circ} - \frac{\angle ADB}{2} = \angle I_2DI, \angle DI_2I_1 = 90^{\circ} + \frac{\angle DCB}{2} = \angle  I_1BI \Rightarrow \triangle DII_2 \sim \triangle I_1IB$ \\

$\Rightarrow \frac{EF}{EI_2} = \frac{BI}{II_2} = \frac{BI_1}{DI_2}$ \\

$\angle ET_3F = \angle EII_2 = \angle I_2DB, \angle EFT_3 = \angle EBC = \angle I_2BD \Rightarrow \triangle ET_3F \sim \triangle I_2DB$ \\

$\Rightarrow \frac{ET_3}{DI_2} = \frac{EF}{BI_2}$ \\

$\Rightarrow  EX = ET_3$ \\

$\angle T_3EI = \angle T_3BI = 90^{\circ} - \angle I_1BE = \angle(EI,IB_1) = \angle IEX$
$\Rightarrow \triangle ET_3I \simeq \triangle EXI$ \\

$\Rightarrow  IX = IT_3$ \\

\centering
\includegraphics[width=0.6\columnwidth]{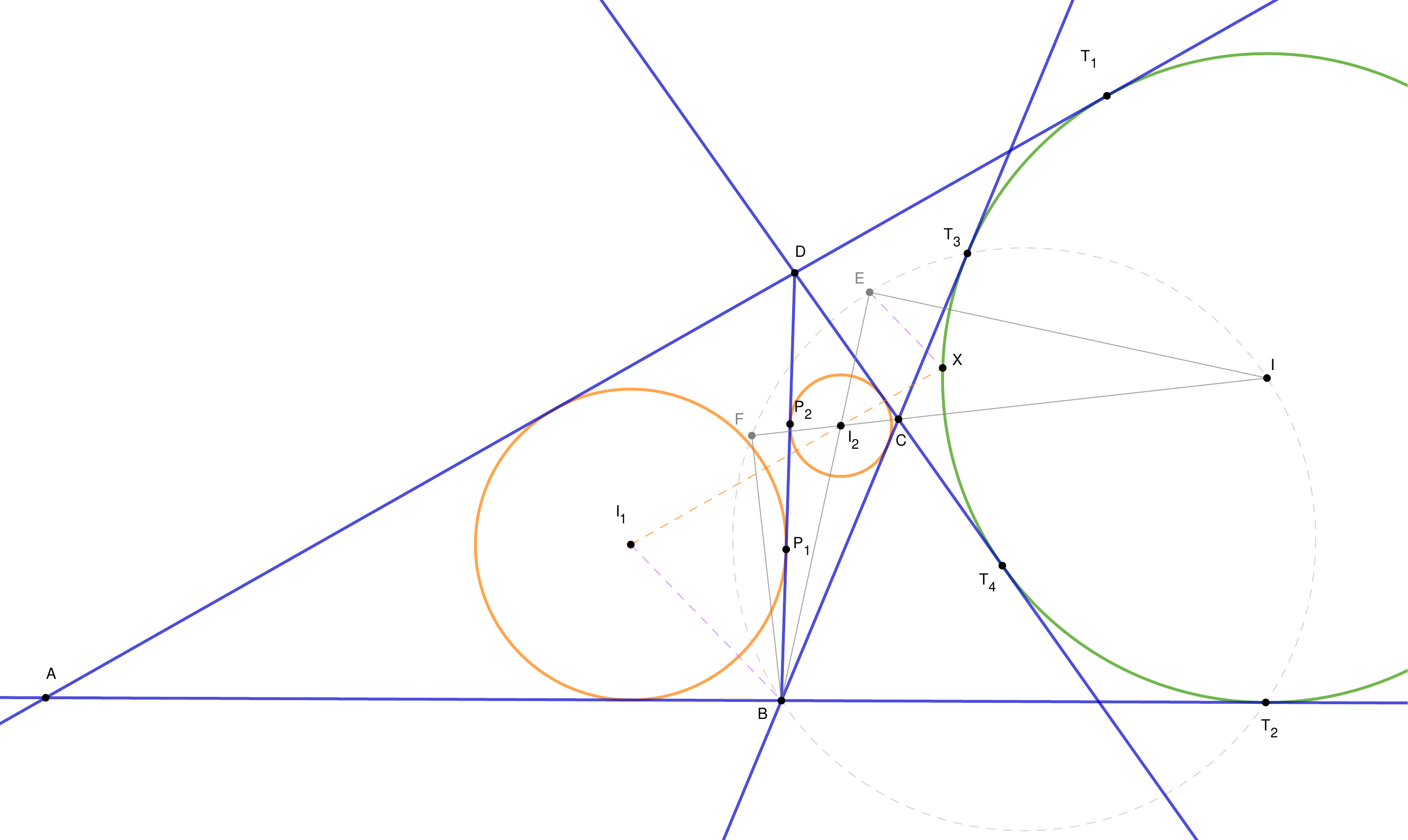}
\label{fig:2008P6}

\end{tcolorbox}
\end{figure}

\clearpage
\section{Prompt to Convert Natural Language into AlphaGeometry's Geometric Language}
\begin{lstlisting}[caption={An illustration of prompt converting a natural language problem into AlphaGeometry's geometry specific language.},
label={prompt:alphageometry},captionpos=b,basicstyle=\ttfamily\scriptsize\linespread{0.9}\selectfont]
Translate a geometry problem from natural language into the following format:



The geometric objects in the converted problem format only contain points.

The naming conventions follows the following rule:

Points are named using lower letters like a,b,c,...,x,y,z, or lower letters with subscripts numbers like a1, b1, a2, ...

Convert the name if not in this convention, for example, convert I_A to i_1



The format is divided into two parts: 1. Problem Definition and 2. Conclusion to be Proved.

Note that the definitions of points are separated by '; ' and definition and conclusion are separated by ' ? '



Problem Definition:

The general structure starts by setting some initial structure and then adding one or more geometry objects each time.



Initial or free structures can be point, segment, triangle, quadrilateral, pentagon, ...

For example:

"Free point A" convert to "a = free"

"Segment AB" convert to "a b = segment"

"Triangle ABC" or "Acute triangle ABC" or "Obtuse triangle ABC" convert to "a b c= triangle".

"Isosceles triangle ABC with AB=AC" convert to "a b c = iso_triangle"

"Right triangle ABC with angle ∠A=90" convert to "a b c = r_triangle".

"Right Isosceles triangle ABC" convert to "a b c = risos"

"Equilateral triangle ABC" "a b c = ieq_triangle"

"Quadrilateral ABCD" is converted to "a b c d = quadrangle".

"Trapezoid ABCD with AB//CD" convert to "a b c d = trapezoid"

"Isosceles Trapezoid ABCD with AB//CD and AD=BC" convert to "a b c d = eq_trapezoid"

"Right trapezoid ABCD with ∠A=90" convert to "a b c d = r_trapezoid"

"Rectangle ABCD" convert to "a b c d = rectangle"

"Pentagon ABCDE" is converted to "a b c d e = pentagon".


Each time a new point is constructed, using one or two constraints,

The constraints could be:

"M is the midpoint of AB" convert to "m = midpoint a b".

"X is foot of A on BC" is converted to "x = foot a b c".

"X as the reflection of A wrt point B" convert to "x = mirror a b".

"X as the reflection of A wrt line BC" convert to "x = reflect a b c".

"X Y are tangent points of the tangent lines from point A to circle centered O passing through B" convert to "x y = tangent x y a o b"

"X satisfies AX is the diameter in (O)" convert to "x = mirror a o"

"The common tangent line l1, l2 of (O1,A) and (O2,B) tangent the two circles at X1,Y1, X2,Y2" convert to "x1 y1 x2 y2 = cc_tangent x y o1 a o2 b"

"O is circumcenter of ABC" convert to "o = circumcenter a b c".

"I is incenter of ABC" convert to "i = incenter a b c".

"H is orthocenter of ABC" convert to "h = orthocenter A B C".

"I is the excenter of ∠A in ABC" convert to "i = excenter a b c",

"I as the excenter of ∠B in ABC" convert to "i = excenter b c a",

"I is the excenter of ∠C in ABC" convert to "i = excenter c a b".

"G is centroid of ABC" convert to "g = centroid a b c"

"N is the nine-point center of ABC" convert to "n = ninepoints a b c".

"X satisfies ABCX is a parallelogram" convert to "x = parallelogram a b c x".

"X Y are points such that XYAB is square" convert to "x y = square a b".

"X satisfies XBC is a equilateral triangle" convert to "x = eq_triangle x b c"

"X such that XA=BC" convert to "x = eqdistance x a b c"

"X on line BC" convert to "x = on_line b c".

"X on parallel line from A to BC" convert to "x = on_pline a b c"

"X on the perpendicular line from A to BC" convert to "x = on_tline a b c"

"X on perpendicular bisector of AB" convert to "x = on_bline a b"

"X on angle bisector of ∠BAC" convert to "x = angle_bisector b a c"

"X on circle with center O pass through A" convert to "x = on_circle o a"

"X on circle passing through A B C" convert to "x = on_circum a b c"

"X on circle with diameter AB" convert to "x = on_dia a b".

"X on tangent from point A to circle centered O passing through A" convert to "x = on_tline a o a".

"X satisfies ∠XAB=∠DEF" convert to "on_aline x a b d e f"

"X satisfies ∠AXB=∠DEF" convert to "eqangle3 x a b e d f"


If X satisfies two constraints, separate the two constraints with ", "

For example, X is the intersection of two conditions, it can be written as "X = condition1, condition2":

"X is the intersection of AB and CD" convert to "X = on_line a b, on_line c d".

"X is the intersection of circles (O, A) and (P, B)" convert to "x = on_circle o a, on_circle p b".

"X is the other intersection of line AB with circle (O, P)" convert to "X = on_line a b, on_circle o p".


Conclusion to be Proved:

"A B C D are cyclic" converted to "? cyclic a b c d".

"A, B, and C are collinear" convert to "? coll a b c".

"AB perpendicular to CD" convert to "? perp a b c d".

"AB parallel to CD" convert to "? para a b c d".

"AB = CD" convert to "? cong a b c d",

"Angle ∠ABC = ∠DEF" convert to "? eqangle a b b c d e e f".

"Ratio AB/CD = EF/GH" convert to "? eqratio a b c d e f g h"

"M is the midpoint of AB" convert to "? midp m a b".

"ABC ~ DEF" or "ABC is similar to DEF", convert to "? simtri a b c d e f".

"ABC ~= DEF" or "ABC is congruent to DEF", convert to "? contri a b c d e f".


Some conclusion need definition of new points:

"AB bisect CD" or "AB pass through the midpoint of CD" convert to "m = midpoint a b ? midp m c d"

"AB, CD, EF are concurrent" convert to "x = on_line a b, on_line c d ? coll x e f".


Note that each new point is separated by a semicolon ";", and the conclusion is separated by " ? ".

Note that newly added points must have different names from previously defined points.


Some conversions require slight modifications:

In triangle ABC, ∠A is that same as ∠BAC.

The incircle or excircle (I) touching BC at point D can be viewed as the foot I on BC, hence convert to "d = foot i b c".

The euler line is the line through orthocenter and circumcenter, X on euler line of ABC convert to "a b c = triangle; o = circumcenter a b c; h = orthocenter a b c; x = on_line o h"

The initial structure (O1),(O2) intersect in A, B we can first define A B then define O1,O2 on perpendicular bisector "a b = segment; o1 = on_bline a b; o2 = on_bline a b"

The initial structure parallelogram ABCD, we need first define ABC then D "a b c = triangle; d = parallelogram a b c d;"

The initial structure cyclic ABCD, we need first define ABC then D "a b c = triangle; d = on_circum a b c;"


Here are a few examples:

Problem 1:

Let $ABCD$ be a trapezoid with $AB\parallel CD$ and $ \Omega $ is a circle passing through $A,B,C,D$. Let $ \omega $ be the circle passing through $C,D$ and intersecting with $CA,CB$ at $A_1$, $B_1$ respectively. $A_2$ and $B_2$ are the points symmetric to $A_1$ and $B_1$ respectively, with respect to the midpoints of $CA$ and $CB$. Prove that the points $A,B,A_2,B_2$ are concyclic.

Convert to:

a b c d = eq_trapezoid; o = on_bline c d; a1 = on_line a c, on_circle o c; b1 = on_line b c, on_circle o c; m1 = midpoint c a; m2 = midpoint c b; a2 = mirror a1 m1; b2 = mirror b1 m2 ? cyclic a b a2 b2


Problem 2:

Given is an acute triangle $ABC$. Its heights $BB_1$ and $CC_1$ are extended past points $B_1$ and $C_1$. On these extensions, points $P$ and $Q$ are chosen, such that angle $PAQ$ is right. Let $AF$ be a height of triangle $APQ$. Prove that angle $BFC$ is a right angle.

Convert to:

a b c = triangle; b1 = foot b a c; c1 = foot c a b; p = on_line b b1; q = on_line c c1, on_tline a a p; f = foot a p q ? perp b f f c


Problem 3:

Let $ABC$ be a triangle. Circle $\Gamma$ passes through $A$, meets segments $AB$ and $AC$ again at points $D$ and $E$ respectively, and intersects segment $BC$ at $F$ and $G$ such that $F$ lies between $B$ and $G$. The tangent to circle $BDF$ at $F$ and the tangent to circle $CEG$ at $G$ meet at point $T$. Suppose that points $A$ and $T$ are distinct. Prove that line $AT$ is parallel to $BC$.

Convert to:

a b c = triangle; o = free; d = on_line a b, on_circle o a; e = on_line a c, on_circle o a; f = on_line b c, on_circle o a; g = on_line b c, on_circle o a; o1 = circumcenter b d f; o2 = circumcenter c e g; t = on_tline f o1 f, on_tline g o2 g ? para a t b c


Problem 4:

Let $ABC$ be a triangle with circumcentre $O$. The points $D,E,F$ lie in the interiors of the sides $BC,CA,AB$ respectively, such that $DE$ is perpendicular to $CO$ and $DF$ is perpendicular to $BO$. (By interior we mean, for example, that the point $D$ lies on the line $BC$ and $D$ is between $B$ and $C$ on that line.) Let $K$ be the circumcentre of triangle $AFE$. Prove that the lines $DK$ and $BC$ are perpendicular.

Convert to:

a b c = triangle; o = circumcenter a b c; d = on_line b c; e = on_line a c, on_tline d c o; f = on_line a b, on_tline d b o; k = circumcenter a e f ? perp d k b c


Problem 5:

In a triangle $ABC$ with a right angle at $C$, the angle bisector $AL$ (where $L$ is on segment $BC$) intersects the altitude $CH$ at point $K$. The bisector of angle $BCH$ intersects segment $AB$ at point $M$. Prove that $CK=ML$

Convert to:

c a b = r_triangle; l = on_line b c, angle_bisector b a c; h = foot c a b; k = on_line c h, angle_bisector b a c; m = on_line a b, angle_bisector b c h ? cong c k m l


Problem 6:

Let $ABC$ be a triangle with $\angle BAC \neq 90^{\circ}.$ Let $O$ be the circumcenter of the triangle $ABC$ and $\Gamma$ be the circumcircle of the triangle $BOC.$ Suppose that $\Gamma$ intersects the line segment $AB$ at $P$ different from $B$, and the line segment $AC$ at $Q$ different from $C.$ Let $ON$ be the diameter of the circle $\Gamma.$ Prove that the quadrilateral $APNQ$ is a parallelogram.

Convert to:

a b c = triangle; o = circumcenter a b c; o1 = circumcenter o b c; p = on_line a b, on_circle o1 o; q = on_line a c, on_circle o1 o; n = mirror o o1 ? para a p n q


Problem 7:

In an acute-angled triangle $ABC$ on the sides $AB$, $BC$, $AC$ the points $H$, $L$, $K$ so that $CH \perp AB$, $HL \parallel AC$, $HK \parallel BC$. Let $P$ and $Q$ feet of altitudes of a triangle $HBL$, drawn from the vertices $H$ and $B$ respectively. Prove that the feet of the altitudes of the triangle $AKH$, drawn from the vertices $A$ and $H$ lie on the line $PQ$.

Convert to:

a b c = triangle; h = foot c a b; l = on_line b c, on_pline h a c; k = on_line a c, on_pline h b c; p = foot h b l; q = foot b h l; x = foot a h k; y = foot h a k ? coll p q x


Problem 8:

Let $ABCD$ be a parallelogram such that $|AB| > |BC|$. Let $O$ be a point on the line $CD$ such that $|OB| = |OD|$. Let $\omega$ be a circle with center $O$ and radius $|OC|$. If $T$ is the second intersection of $\omega$ and $CD$, prove that $AT, BO$ and $\omega$ are concurrent.

Convert to:

a b c = triangle; d = parallelogram a b c d; o = on_line c d, on_bline b d; t = on_line c d, on_circle o c; x = on_line a t, on_line o b ? cong o x o c


Problem 9:

As shown in the figure below, the in-circle of $ABC$ is tangent to sides $AB$ and $AC$ at $D$ and $E$ respectively, and $O$ is the circumcenter of $BCI$. Prove that $\angle ODB = \angle OEC$.

Convert to:

a b c = triangle; i = incenter a b c; d = foot i a b; e = foot i a c; o = circumcenter b c i ? eqangle o d d b c e e o


Problem 10:

In acute triangle $ABC,$ $AB<AC,$ $O$ is the circumcenter of the triangle. $M$ is the midpoint of segment $BC,$ $(AOM)$ intersects the line $AB$ again at $D$ and intersects the segment $AC$ at $E.$ Prove that $DM=EC.$

Convert to:

a b c = triangle; o = circumcenter a b c; m = midpoint b c; o1 = circumcenter a o m; d = on_line a b, on_circle o1 a; e = on_line a c, on_circle o1 a ? cong d m e c


Please follow these examples to convert the following problems:
{INPUT_PROBLEM}
\end{lstlisting}

\clearpage
\section{Prompt to Convert Natural Language into our Geometric language}
\begin{lstlisting}[caption={An illustration of prompt converting a natural language problem into our geometry specific language.},
label={prompt:sigmageometry},captionpos=b,basicstyle=\ttfamily\scriptsize\linespread{0.9}\selectfont]
Translate a geometry problem from natural language into the following format:

The geometric objects in the problem include points, lines, and circles. 

The naming conventions follows the following rule:
Points are named using uppercase letters like A, B, C...X, Y, Z, or uppercase letters with subscripts (numbers or lowercase letters) like A1, B1, Ai, Bj, I_1,I_a...

Convert the name if not in this convention, for example, convert I_A to I_a
Lines are named using lowercase letters like l, m... or lowercase letters with numbers like l1, m2,... or by the names of two points on the line like AB, CD.

Circles are named using the center of the circle of the circle (O), (P),... or diameter of the circle (AB),(CD),... or three points on the circle with parentheses  or or (ABC), (DEF),... or the center and a point on the circle separated by a comma (O, A), (O, P)... or similar with lines with lower case letter  c,c1,... or greek letter such as Γ,γ,Ω,ω...

The format is divided into two parts: 1. Problem Definition and 2. Conclusion to be Proved.

Problem Definition:

The general structure starts by setting some initial structure and then adding one or more geometry objects each time.

Initial structures can be triangle, acute_triangle, or right_triangle, quadrilateral, pentagon, ... 
or point line circle:

"Triangle ABC" convert to "A B C = triangle".

"Acute triangle ABC" convert to "A B C = acute_triangle".

"Right triangle ABC" convert to "A B C = right_triangle".

"Obtuse triangle ABC" convert to "A B C = obtuse_triangle".

"Isosceles triangle ABC with AB=BC" convert to "B A C = isos_triangle"

"Equilateral triangle ABC" "A B C = equilateral_triangle"

"Quadrilateral ABCD" is converted to "A B C D = quadrilateral".

"Cyclic quadrilateral ABCD" is converted to "A B C D = cyclic_quadrilateral".

"Tangential quadrilateral ABCD" convert to "A B C D = tangential_quadrilateral".

"Pentagon ABCDE" is converted to "A B C D E = pentagon".

"Hexagon ABCDEF" is converted to "A B C D E F = hexagon".


Each time a new point or line or circle is constructed, the definition for each point is given using one or two constraints,

The constraints could be:

"A on line BC" convert to "A = on_line BC".

"A on circle (O)" convert to "A = on_circle (O)".

"D is the foot of A on BC" is converted to "D = foot A BC".

"M is the midpoint of AB" convert to "M = midpoint A B".

"M is the midpoint of arc ABC" convert to "M = midarc A B C"

"M is the midpoint of arc AB without C" convert to "M = midarc_no A B C"

"O is the circumcenter of ABC" convert to "O = circumcenter ABC".

"H is the orthocenter of ABC" convert to "H = orthocenter A B C".

"I is the incenter of ABC" convert to "I = incenter A B C".

"N is the nine-point center of ABC" convert to "N = ninepointcenter A B C".

"I1 is the excenter relative to  ∠A in ABC" convert to "I1 = excenter A B C", 

"I2 as the excenter relative to  ∠B in ABC" convert to "I2 = excenter B C A",

"I3 is the excenter relative to  ∠C in ABC" convert to "I3 = excenter C A B".

"P is the isogonal conjugate of Q wrt ABC" convert to "P = isogonal_conjugate A B C".

"X satisfies ABCX is a parallelogram" convert to "X = parallelogram A B C X".

"X Y are points such that XYAB is square" convert to "X Y = square A B".

"X as the reflection of A wrt line BC" convert to "X = reflect A BC".

"X as the reflection of A wrt point B" convert to "X = reflect A B".

"The radical axis of circles (A) and (B)" convert to "radical_axis (A) (B)".

"Tangent from point A to circle (O)" convert to "tangent_line A (O)".

"the tangent line from A to (O) intersect (O) at X, Y" convert to "X Y = tangent_point A (O)"

"Euler line of triangle ABC" convert to "euler_line A B C".

"perpendicular bisector of AB" convert to "perpendicular_bisector AB".

"The angle bisector of angle ABC" or "The angle bisector of angle B in triangle ABC" convert to  "angle_bisector A B C".

"The external angle bisector of angle ABC" or "The external angle bisector of angle B in triangle  ABC" convert to "angle_exbisector A B C".

"line through A parallel to BC" convert to "parallel_line A BC".

"perpendicular line through A to BC" convert to "perpendicular_line A BC".

"circle with diameter AB" convert to "(AB)".

"X satisfies AX=BC" convert to "length_equal A X B C"

"X satisfies <AXY=<PQR" convert to "angle_equal A X Y P Q R"

"O is the circumcenter of triangle formed by lines l1,l2,l3" convert to "X Y Z = triangle l1 l2 l3;  O = circumcenter X Y Z" here X Y Z should be new points with dfferent names

"AB is the diameter in (O)" (where A is already defined) convert to "B = antipodal A (O)"

"l is a line through A" convert to "l = line_through A"

"ω is a circle through A B" convert to "ω = circle_through A B"


If X satisfies two constraints, separate the two constraints with ", "
For example, X is the intersection point under two conditions, it can be written as "X = condition1, condition2" or "X = intersection condition1, condition2" . :


"X is the intersection of AB and CD" convert to "X = AB, CD".

"X is the intersection of circles (O, OA) and (P, PB)" convert to "X = (O, OA), (P, PB)".

"X is the other intersection of line AB with circle (O, OP)" convert to "X = AB, (O, OP)".

"X as the other intersection of line PQ with circle (ABC)" convert to "X = PQ, (ABC)".

Note that there should always be a "," between two conditions.

Conclusion to be Proved:

"A B C D are cyclic" converted to "concyclic A B C D".

"A, B, and C are collinear" convert to "collinear A B C".

"AB, CD, EF are concurrent" convert to "concurrent AB CD EF".

"AB perpendicular to CD" convert to "perpendicular AB CD".

"AB parallel to CD" convert to "parallel AB CD".

"AB = CD" convert to "cong AB CD", 

"∠ABC = ∠DEF" convert to "equal_angle ABC DEF".

"M is the midpoint of AB" convert to "midpoint M A B".

"ABC ~ DEF" or "ABC is similar to DEF", convert to "similar ABC DEF".

"ABC ~= DEF" or "ABC is congruent to DEF", convert to "contri ABC DEF".

"ABCD is a square" convert to "square A B C D".

"Circles (O1), (O2), and (O3) having a common radical axis" or "(O1), (O2), (O3) are coaxal" convert to "coaxal (O1) (O2) (O3)".

"AB bisect CD" or "AB pass through the midpoint of CD" convert to "bisect AB CD" 


Note that each new point is separated by a semicolon ";", and the conclusion is separated by "Prove: ".

Note that newly added points must have different names from previously defined points.


Some conversions require slight modifications:

In triangle ABC, ∠A is that same as ∠BAC.

The incircle or excircle (I) touching BC at point D can be viewed as the foot I on BC, hence convert to "D = foot I BC" or directly "D = (I), BC".


Note that AB represents a line or segment without any space in between, such as in "concurrent AB CD EF" or "foot X AB".

Here are a few examples:

Problem 1: 

In a cyclic hexagon ABCDEF, extend AB and CD to intersect at point G. Extend AF and DE to intersect at point H. Let M and N be the circumcenters of triangles BCG and EFH, respectively. Prove that lines BE, CF, and MN are concurrent.

Convert to:

A B C D E F = cyclic_hexagon; G = AB, CD; H = AF, DE; M = circumcenter B C G; N = circumcenter E F H; Prove: concurrent BE CF MN


Problem 2: 

Let H be the orthocenter of triangle ABC, and P be any point. The lines AP, BP, and CP intersect the circumcircle of triangle ABC at points A1, B1, and C1, respectively. Let A2, B2, and C2 be the reflections of A1, B1, and C1 over the sides BC, CA, and AB, respectively. Then, A1, B1, C1, and H are concyclic.

Convert to:

A B C = triangle; H = orthocenter A B C; P = point; A1 = AP, (ABC); B1 = BP, (ABC); C1 = CP, (ABC); A2 = reflect A1 BC; B2 = reflect B1 CA; C2 = reflect C1 AB; Prove: concyclic A1 B1 C1 H


Problem 3: 

In triangle ABC, the incircle (I) touches sides BC, CA, and AB at points D, E, and F, respectively. Prove that lines AD, BE, and CF are concurrent.

Convert to:

A B C = triangle; I = incenter A B C; D = foot I BC; E = foot I CA; F = foot I AB; Prove: concurrent AD BE CF


Problem 4:

Let $ABC$ be a triangle with $AC > BC,$ let $\omega$ be the circumcircle of $\triangle ABC,$ and let $r$ be its radius. Point $P$ is chosen on $\overline{AC}$ such taht $BC=CP,$ and point $S$ is the foot of the perpendicular from $P$ to $\overline{AB}$. Ray $BP$ mets $\omega$ again at $D$. Point $Q$ is chosen on line $SP$ such that $PQ = r$ and $S,P,Q$ lie on a line in that order. Finally, let $E$ be a point satisfying $\overline{AE} \perp \overline{CQ}$ and $\overline{BE} \perp \overline{DQ}$. Prove that $E$ lies on $\omega$. 

Convert to:

A B C = triangle; O = circumcenter A B C; P = AC, (C,B); S = foot P AB; D = intersection BP, (ABC); Q = SP, (P,OA); l1 = perpendicular_line A CQ; l2 = perpendicular_line B DQ; E = intersection l1, l2; Prove: on_circle E (ABC)


Problem 5:

Let $ABC$ be an acute, scalene triangle with orthocentre $H$. Let $\ell_a$ be the line through the reflection of $B$ with respect to $CH$ and the reflection of $C$ with respect to $BH$. Lines $\ell_b$ and $\ell_c$ are defined similarly. Suppose lines $\ell_a$, $\ell_b$, and $\ell_c$ determine a triangle $\mathcal T$. Prove that the orthocentre of $\mathcal T$, the circumcentre of $\mathcal T$, and $H$ are collinear. 

Convert to:

A B C = acute_triangle; H = orthocenter A B C; B1 = reflect B CH; C1 = reflect C BH; la = line B1 C1; A1 = reflect A BH; B2 = reflect B AH; lb = line A1 B2; A2 = reflect A CH; C2 = reflect C AH; lc = line A2 C2; X = lb, lc; Y = la, lc; Z = la, lb; H1 = orthocenter X Y Z; O1 = circumcenter X Y Z; 
Prove: collinear H H1 O1

Problem 6:

Let $ABCD$ be a parallelogram with $AC=BC.$ A point $P$ is chosen on the extension of ray $AB$ past $B.$ The circumcircle of $ACD$ meets the segment $PD$ again at $Q.$ The circumcircle of triangle $APQ$ meets the segment $PC$ at $R.$ Prove that lines $CD,AQ,BR$ are concurrent. 

Convert to:

C A B = isos_triangle; D = parallelogram A B C D; P = on_line AB; Q = intersection PD, (ACD); R = intersection PC, (APQ); Prove: concurrent CD AQ BR


Problem 7:

Let $I$ be the incentre of acute triangle $ABC$ with $AB\neq AC$. The incircle $\omega$ of $ABC$ is tangent to sides $BC, CA$, and $AB$ at $D, E,$ and $F$, respectively. The line through $D$ perpendicular to $EF$ meets $\omega$ at $R$. Line $AR$ meets $\omega$ again at $P$. The circumcircles of triangle $PCE$ and $PBF$ meet again at $Q$. Prove that lines $DI$ and $PQ$ meet on the line through $A$ perpendicular to $AI$. 

Convert to:

A B C = acute_triangle; I = incenter A B C; D = foot I BC; E = foot I CA; F = foot I AB; R = intersection perpendicular_line D EF, (I); P = intersection AR, (I); Q = intersection (PCE), (PBF); X = DI, PQ; Prove: perpendicular AI AX


Problem 8:

Let $ABC$ be an acute triangle with circumcircle $\Gamma$. Let $\ell$ be a tangent line to $\Gamma$, and let $\ell_a, \ell_b$ and $\ell_c$ be the lines obtained by reflecting $\ell$ in the lines $BC$, $CA$ and $AB$, respectively. Show that the circumcircle of the triangle determined by the lines $\ell_a, \ell_b$ and $\ell_c$ is tangent to the circle $\Gamma$. 

Convert to:

A B C = acute_triangle; ω = circumcircle A B C; X = on_circle ω; l = tangent X ω; la = reflect l BC; lb = reflect l CA; lc = reflect l AB; Ω = circumcircle la lb lc; Prove: tangent ω Ω


Problem 9:

Let $\Omega$ and $O$ be the circumcircle and the circumcentre of an acute-angled triangle $ABC$ with $AB > BC$. The angle bisector of $\angle ABC$ intersects $\Omega$ at $M \ne B$. Let $\Gamma$ be the circle with diameter $BM$. The angle bisectors of $\angle AOB$ and $\angle BOC$ intersect $\Gamma$ at points $P$ and $Q,$ respectively. The point $R$ is chosen on the line $P Q$ so that $BR = MR$. Prove that $BR\parallel AC$. 

Convert to:

A B C = acute_triangle; O = circumcenter A B C; M = midarc_no A C B; Γ = circle (BM); P = intersection perpendicular_line O AB, Γ; Q = intersection perpendicular_line O BC, Γ; R = PQ, length_equal B R M R; Prove: parallel BR AC


Problem 10:

Let $ABC$ be a triangle with $\angle B > \angle C$. Let $P$ and $Q$ be two different points on line $AC$ such that $\angle PBA = \angle QBA = \angle ACB $ and $A$ is located between $P$ and $C$. Suppose that there exists an interior point $D$ of segment $BQ$ for which $PD=PB$. Let the ray $AD$ intersect the circle $ABC$ at $R \neq A$. Prove that $QB = QR$. 

Convert to:

A B C = triangle; P = AC, angle_equal P B A B C A; Q = AC, angle_equal Q B A A C B; D = BQ, length_equal P D P B; R = intersection AD, (ABC); Prove: cong Q B Q R



Please follow these examples to convert the following problem:
{INPUT_PROBLEM}
\end{lstlisting}

\end{document}